# Three Types of Negation of Triple and its Elements and an Extension of Triple

Zhenghua Pan

School of Mathematics and Data Science, Jiangnan University, Wuxi 214122, China

panzh@jiangnan.edu.cn

**Abstract:** In various data models, the classical triple <s, p, o> is a typical semantic data model. However, due to the design of the triple as a simple structure for representing positive assertions, it cannot sufficiently express different forms of negation present in the triple and its elements. This paper conceptually proposes that there are three distinct forms of negation within triples and their elements: contradictory negation, opposite negation and intermediary negation. Based on the the set SCOI and the logic LCOI+PLCOI with three kinds of negation, we propose an extension of triple that can distinguish and express these three different negations in the triple and its elements, called the "TCOI triple with contradictory negation, opposite negation and intermediary negation".

The TCOI triple is a semantic and structural extension of the classical triple. While retaining the ability to express positive assertions, it systematically introduces the three semantic dimensions of contradictory negation, opposite negation and intermediary negation, allowing these negations to independently act on the elements (s, p, o) of the triple and on the whole triple. This significantly enhances the triple model capability to represent and reasoning about complex negative information.

This paper also explores the expressive power and reasoning of the TCOI triple. Based on the semantics of the logic LCOI+PLCOI, we introduce the notion of "TCOI-entailment" as the semantics implication in TCOI triple implication reasoning, thereby establishing a connection between TCOI triple implication reasoning and inferences in the logic LCOI+PLCOI. This shows that LCOI+PLCOI provide the logical foundation for TCOI triple implication reasoning. Furthermore, we discuss the application of TCOI triple implication reasoning in counterfactuals and counterfactual reasoning involving the three different types of negation. We propose a truth-value (continuous value) algorithm for TCOI triple implication reasoning and perform its calculation through an example of the counterfactuals and counterfactual reasoning.

This work was supported by the National Natural Science Foundation of China [60575038, 60973156, 61375004]; the State Key Laboratory for Novel Software Technology, Nanjing University, P.R. China [KFKT2020B01]

# 1. Introduction

In various data models, triples are a typical semantic data model. They express semantic information in an ordered format <subject, predicate, object> (or <s, p, o>). Triples play an irreplaceable and unique role in knowledge representation, knowledge reasoning, and knowledge interconnection in a concise manner. The most profound significance of triples is that they provide machines with a standardized language to simulate human expression of “facts” and “relationships” [1]. Triples are widely used in fields such as the Semantic Web, Resource Description Framework (RDF), knowledge graphs, graph computing, information retrieval and semantic search, natural language processing (NLP), as well as artificial intelligence and machine learning [2-8].

Negation is a complex natural language phenomenon and also an important basic concept in knowledge [9]. Broadly speaking, negation connects an expression E with another expression whose meaning is, in some sense, opposed to that of E [10]. Therefore, the key challenge in understanding negation is to identify the meaning that is in some way opposed to E—this is a semantically complex and highly vague task [11]. Negation in triples is of significant importance in knowledge representation and reasoning. It can be said that negation in triples is an indispensable "other half" of knowledge representation and reasoning. It allows systems (such as Resource Description Framework and knowledge graphs) to extend from merely describing what exists to being able to articulate what does not exist, thereby achieving logical completeness, precision of knowledge, and depth of reasoning.

Regarding the negativity of triples, from the semantic level of the triple structure, there are two forms: one is negativity at the triple level, and the other is negativity at the element level. These two forms of negativity do not have only one form of negation in practice, but rather different types of negation with different meanings.

For example: in triple expressions, for the triple T: <*x*, property, positive number> (i.e., the statement “*x* is a positive number”), the triples T1: <*x*, property, non-positive number> (i.e., the statement “*x* is not a positive number”), T2: <*x*, property, negative number > (i.e., the statement “*x* is a negative number”) and T3: <*x*, property, zero> (i.e., the statement “*x* is zero”) represent different negations of triple T. In the expression of elements within triple, for the predicate “likes” in the triple <he, likes, her>, ‘dislikes’, ‘hates’ and ‘neither likes nor hates’ are three different types of negation for the predicate “likes”.

For the different negations present in triples and their elements, the classical triple <s, p, o>, as a structure expressing simple positive semantic relationships, can only express affirmative assertions between the subject, predicate, and object, lacking an inherent mechanism to distinguish and express these different types of negation.

For the classical triple <s, p, o>, since there is only one type of negation (classical negation) in the formal languages of classical and non-classical logic, it cannot distinguish or express these different forms of negation existing in triples and their elements. Therefore, to accurately and precisely express these different negations in knowledge representation, the classical triple <s, p, o> needs to be extended based on a non-classical logic that can distinguish and express different negations at both the syntactic and semantic interpretation levels. Such logic includes multiple negation operators or semantic clarification mechanisms to achieve complete expression of negation meanings and accurate semantic modeling.

In this paper, we conceptually propose that there exist three different forms of negation within triple and their elements: “contradictory negation”, “opposite negation”, and “intermediary negation”. Based on the set SCOI and the logic LCOI+PLCOI with three kinds of negation [12, 13], we present an extension of triple capable of distinguishing and expressing these negations: the “TCOI triple with contradictory negation, opposite negation and intermediary negation”. TCOI triple is a semantic and structural extension of the classical triple <s, p, o>. While retaining the ability to express affirmative assertions, it systematically introduces the three semantic dimensions of contradictory negation, opposite negation and intermediary negation, allowing these negations to independently apply to the elements (s, p, o) of the triple and on the whole triple. This significantly enhances the triple model capability to represent and reasoning about complex negative information.

In this paper, we also discuss the expressive power and reasoning of the TCOI triple. Based on the semantics of the logic LCOI+PLCOI, we introduce the notion of "TCOI-entailment" as the semantics implication in TCOI triple implication reasoning. Through TCOI-entailment, a connection is established between TCOI triple implication reasoning and inferences in the logic LCOI+PLCOI. This demonstrates that the formal inference properties already proven in LCOI+PLCOI are valid in TCOI triple implication reasoning. LCOI+PLCOI provide the logical foundation for TCOI triple implication reasoning. Furthermore, we discuss the application of TCOI triple entailment reasoning in counterfactuals and counterfactual reasoning involving the three types of negation. We propose a truth-value (continuous value) algorithm for TCOI triple implication reasoning and perform truth-value calculation on an example of counterfactuals and counterfactual reasoning.

As far as we know, there is no direct evidence indicating the existence of research specifically focusing on negation within triples and their elements. This paper investigates negation in triples and proposes that there exist three different types of negation within triples and their elements, as well as an extension of triples capable of distinguishing and expressing these negations—work that, to date, has not been seen. The main contributions of this paper are as follows:

1. Conceptually proposing that three different types of negation exist within triple and their elements: contradictory negation, opposite negation, and intermediary negation.
2. Under the set SCOI and the logic LCOI+PLCOI framework, we propose a triple that can express three different types of negations present in triple and their elements: "TCOI triple with contradictory negation, opposite negation and intermediary negation". TCOI not only distinguishes between clear triple and fuzzy triple, as well as the correlation degree between the subject and the object, but it is also capable of expressing the contradictory negation, opposite negation and intermediary negation for the triple and its elements (predicate, object).
3. Based on the semantics of the logic LCOI+PLCOI, we introduce a notion "TCOI-entailment" as the implication in TCOI triple implication reasoning. Through TCOI-entailment, we establish a connection between TCOI triple implication reasoning and inferences in the logic LCOI+PLCOI. This shows that the formal inference properties already proven in LCOI+PLCOI are valid in TCOI triple implication reasoning. LCOI+PLCOI provide the logical foundation for TCOI triple implication reasoning.
4. Applying TCOI triple implication reasoning to counterfactuals and counterfactual reasoning. A truth-value (continuous value) algorithm for TCOI triple implication reasoning is proposed, and a calculation is performed on an example of counterfactuals and counterfactual reasoning.

The organization of this paper is as follows. Section 2 discusses related work. Section 3 conceptually discusses the three types of negation and their characteristics within triples and their elements. Section 4 introduces the basic definitions of the logical set SCOI and the logic LCOI+PLCOI with three types of negation. It proposes a continuous truth-value semantics with the truth domain [0, 1] for logic LCOI+PLCOI and discusses its metalogical properties. In Section 5, based on LCOI+PLCOI, the TCOI triple with three types of negation is proposed. Section 6 discusses the expressive power of the TCOI triple. Section 7 explores TCOI triple reasoning and its applications. Section 8 summarizes the main conclusions of this paper and future work.

## 2. Related works

This paper aims to conceptually propose that there exist three different forms of negation in general triple and their elements: contradictory negation, opposite negation, and intermediary negation. Furthermore, based on the set SCOI and the logic LCOI+PLCOI, it proposes an extension of triple that can distinguish and express these negations.

Among related studies, reference [13] is the only one most closely related to this paper. In the extension of the Resource Description Framework called RDFCOI, it proposes and defines a suitable triple for RDFCOI, the RDFCOItriple. The RDFCOI triple maintains the definition of the RDF triple and is capable of expressing the three types of negation for the RDFCOI. However, that work is not specifically focused on the negation of triples and their elements, nor does it discuss the reasoning and logical foundation of triples with three types of negation. Therefore, the purpose and significance of this paper differ from those of reference [13]. The related work relevant to this paper can be divided into two aspects: the study of negation of whole triple, and the study of negation of elements within triple.

Regarding the negation of whole triple, there are no direct studies found in the literature. However, several research works related to triples demonstrate the practical role and significance of negation. Among the most directly relevant papers, one extends RDFS to handle negative statements under the Open World Assumption (OWA) and explicitly avoids reification for negative triples [14]. Multiple negation concepts are operationalized through weak negation, strong negation, local closed world assumption, and scoped negation as failure, explicitly formalizing negation similar to that in RDF knowledge representation [15, 16]. Negative statements are considered useful; one proposal is ERDF, where an RDF triple can be positive or negative and distinguishes between weak and strong negation [17]. Another work directly addresses negative statements and reports that negation via SPARQL MINUS does not yield relevant knowledge in the studied cases [18]. There are broader surveys on negative statements, completeness, and partial closed-world semantics around open-world knowledge bases [19]. Another closely related area is negation in queries. Reference [20] analyzes SPARQL negation operators, their expressive power, and algebraic behavior. SPARQL property path negation is related to finite negation on graph navigation patterns rather than simple assertion triples [21, 22]. Some works explicitly study queries with negation under RDF's open-world setting and define soundness checking for them [23], and so on.

Regarding the negation of elements within triple, there are no explicit analytical studies focusing on the negation of the three elements, nor are there works presenting a comprehensive extended triple that can handle negation of the subject, predicate, and object separately. The most directly relevant research explicitly involves negation in the predicate position, while negation of the subject and object is typically handled indirectly under the broader notions of "negative statements" or "negative knowledge" in RDF/SPARQL [24, 25, 26]. For the negation of predicate, SPARQL triple patterns with negation have been explicitly extended at the predicate position, with their syntax and semantics defined [26]. For the negation of object, modeling of negative statements such as "no treatment" and general negation in relational data frameworks (RDF) has been proposed, introducing a minimal entailment system for RDFS that includes negative statements [24]. Additionally, representation and querying of negative knowledge in RDF cover practical aspects of missing/negated information at the object level [25]. For the negation of subject, research exists on generating negative statements about real-world entities, i.e., subject-centered negation, although this is not formal RDF triple logic [27].

In summary, there is no direct evidence of research specifically addressing negation in triples and their elements, nor is there an extension of triples that can separately handle negation of the triple and its individual elements.

# 3. Three types of negation in triple and its elements and their characteristics

A triple expresses a complete and indivisible statement whose meaning is determined by concepts (the smallest unit in a statement), with the elements of the triple (s, p, o) each representing different concepts. Therefore, the negation of the triple <s, p, o> and its elements can be uniformly reduced at the conceptual level to the negation of atomic concepts (the elements) or the compound of atomic concepts (the whole triple). In other words, the negation of the triple and its elements is based on the negation of atomic concepts.

In references [12] and [13], we distinguish between “crisp concepts” and “fuzzy concepts” at the conceptual level and thoroughly understand the “contradiction” and “opposition” within concepts, thereby proposing that three different types of negation exist in atomic concepts: contradictory negation, opposite negation, and intermediary negation. Therefore, we believe that these three different forms of negation must also exist within triples and their elements.

In this section, based on a brief overview of these three forms of negation, we discuss their characteristics.

(1) *Contradictory Negation*. For a species concept under a genus concept, another species concept that has a contradictory relationship with it constitutes a form of negation. We refer to this type of negation as “contradictory negation”. In this form of negation, the intensions (connotations) of the two species concepts mutually negate each other, the extensions (denotations) are mutually exclusive (either one or the other), and the sum of the extensions equals the extension of the genus concept. For example, for the species concept ‘positive integer’ under the genus concept “integer”, another species concept ‘non-positive integer’ is its contradictory negation. For the two species concepts ‘daytime’ and ‘non-daytime’ under the genus concept “one day”, the latter is the contradictory negation of the former. From this, it can be known that the negation in classical logic is precisely this kind of negation.

(2) *Opposite Negation*. For a species concept under a genus concept, another species concept that has an oppositional relationship with it constitutes another form of negation. We refer to this type of negation as “opposite negation”. In this form of negation, the intensions of the two species concepts mutually negate each other and exhibit the greatest difference in intension, but their extensions are not mutually exclusive (not either-or), and the sum of their extensions is less than the extension of the genus concept. For example, for the species concept ‘positive integer’ under the genus concept “integer”, another species concept ‘negative integer’ is its opposite negation. For the two species concepts ‘daytime’ and ‘night’ under the genus concept “one day”, the latter is the opposite negation of the former.

(3) *Intermediary Negation*. The intermediary concept between opposite concepts constitutes a (weak) form of negation of the opposite concepts. We refer to this type of negation as “intermediary negation”. In this form of negation, the opposite concepts transition through the intermediary concept and the sum of their extensions equals the extension of the genus concept. For example, under the genus concept “integer”, between the two opposing species concepts ‘positive integer’ and ‘negative integer’, the concept “zero” is their intermediary negation. Under the genus concept “one day”, between the two opposing species concepts ‘daytime’ and ‘night’, the concepts “dusk” and “dawn” are their intermediary negation.

From the meanings of the three types of negations mentioned above, the contradictory negation is the traditional negation. The opposite negation can be referred to as strong negation, and intermediary negation as weak negation.

To fully understand the meaning of the above three kinds of negation, we further discuss their characteristics in terms of both the intension of the concepts as well as their extensional relations.

(1) Contradictory Negation in Clear Concepts (CNC)

Characteristics of CNC: Extensions are clear, either this or that, and the sum of extensions is equal to the extension of the genus concept.

For example, the positive integer and non-positive integer under the genus concept of “integer” are clear concepts, while the non-positive integer is the contradictory negation of positive integer. The diagram illustrating the extensional relationship between them is shown below (Figure 1).

| positive integer | non- positive integer |
|---|---|

**Fig. 1** The extensional relationship between positive integer and non-positive integer

(2) Opposite Negation in Clear Concepts (ONC)

Characteristics of ONC: Extensions are clear, not “either this or that”, and the sum of the extensions is less than the extension of the genus concept.

For example, the positive integer and negative integer under the genus concept of “integer” are clear concepts, while the negative integer is the opposite negation of positive integer. The diagram illustrating the extensional relationship between them is shown below (Figure 2).

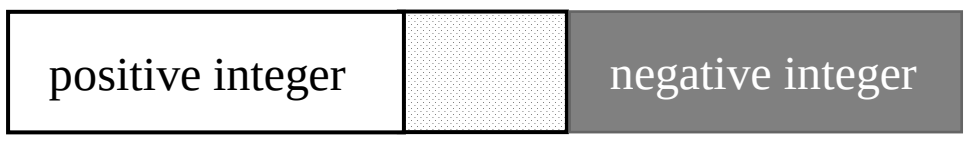

**Fig. 2** The extensional relationship between positive integer and negative integer

(3) Intermediary Negation in Clear Concepts (INC)

Characteristics of INC: Extensions are clear, opposing sides transition to each other through ‘intermediaries’, and the sum of the extensions equals the extension of the genus concept.

For example, zero, positive integer and negative integer are clear concepts under the genus concept of “integer”. Zero serves as an ‘intermediary’ between positive integer and negative integer, and it is the intermediary negation of both positive integer and negative integer. The diagram illustrating the extensional relationship between them is shown below (Figure 3).

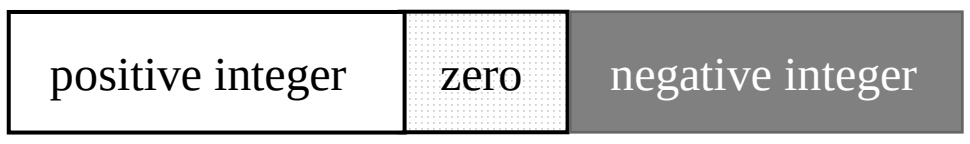

**Fig. 3** The extensional relationship between positive integers, negative integers, and zero.

(4) Contradictory Negation in Fuzzy Concepts (CNF)

Characteristics of CNF: Extensions are not clear, either this or that, and the sum of extensions is equal to the extension of the genus concept.

For example, the daytime and non-daytime under the genus concept of “day” are fuzzy concepts, while the non-daytime is the contradictory negation of daytime. The diagram illustrating the extensional relationship between them is shown below (Figure 4).

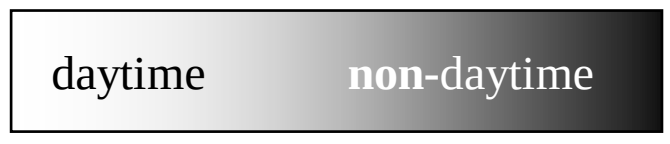

**Fig. 4** The extensional relationship between daytime and non-daytime

(5) Opposite Negation in Fuzzy Concepts (ONF)

Characteristics of ONF: Extensions are not clear, not “either this or that”, and the sum of the extensions is less than the extension of the genus concept.

For example, the daytime and night under the genus concept of “day” are fuzzy concepts, while the night is the opposite negation of daytime. The diagram illustrating the extensional relationship between them is shown below (Figure 5).

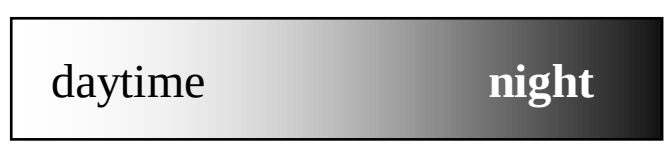

**Fig. 5** The extensional relationship between daytime and night

(6) Intermediary Negation in Fuzzy Concepts (INF)

Characteristics of IFC: Extensions are not clear, opposing sides transition to each other through ‘intermediaries’, and the sum of the extensions equals the extension of the genus concept.

For example, dusk, daytime and night are fuzzy concepts under the genus concept of “day”. Dusk serves as an ‘intermediary’ between daytime and night, and it is the intermediary negation of both daytime and night. The diagram illustrating the extensional relationship between them is shown below (Figure 6).

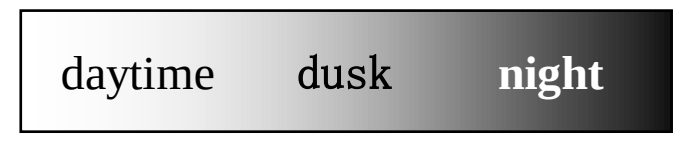

**Fig. 6.** The extensional relationship between daytime, night, and dusk

The above distinction at the conceptual level differentiates contradictory negation, opposite negation, and intermediary negation in atomic concepts, clarifying that negation in concepts is not a single variant of the classical “not”. Since the negation of triple and their elements is logically founded upon the negation of atomic concepts, the three types of negation in atomic concepts must correspondingly be the forms of negation in triples and their elements. They constitute the semantic primitives of the entire knowledge represented by the triple.

# 4. Set and logic with three kinds of negation

For the three kinds of negation present in the aforementioned concepts, in order to establish a mathematical foundation that can fully reflect them along with their properties, relationships, and laws, we proposed a set SCOI and logic LCOI+PLCOI with contradictory negation, opposite negation and intermediary negation that takes clear and fuzzy entities as the research objects [12,13].

In this section, we provide an overview of SCOI and LCOI+PLCOI, and propose a continuous-valued semantics for LCOI+PLCOI with a truth value range of [0, 1]. Under this semantics, the soundness theorem for LCOI+PLCOI is proved.

## 4.1 Set with contradictory negation, opposite negation and intermediary negation

We use symbols $\neg$, ╕, and $\sim$ to represent ‘contradictory negation’, ‘opposite negation’, and ‘intermediary negation’, respectively.

**Definition 1**. Let $U$ be universe of discourse, $\lambda\in(0, 1)$. Mapping $f$: $U \to [0, 1]$ confirms a set $A$ on $U$, call $f$ the membership function of $A$, and $f(x)$ the membership degree of $x$ to $A$ (denoted as $A(x)$).

(1) If $A$ is a fuzzy set, then

($i$). Mapping $f^{╕}: \{A(x) \mid x\in U\} \to [0, 1]$ confirms a fuzzy set $A^{╕}$ on $U$, $A^{╕}(x) = f^{╕}(A(x)) = 1- A(x)$. Call $A^{╕}$ the opposite negation set of $A$.

($ii$). Mapping $f^{\sim}: \{A(x) \mid x\in U\} \to [0, 1]$ confirms a fuzzy set $A^{\sim}$ on $U$, $A^{\sim}(x) = f^{\sim}(A(x))$. Call $A^{\sim}$ the intermediary negation set of $A$. Where

$$A^{\sim}(x) = \begin{cases} \lambda-\dfrac{2\lambda-1}{1-\lambda}(A(x)-\lambda), & \text{when } \lambda\in[½, 1) \text{ and } A(x)\in(\lambda, 1] \quad \text{(a)} \\ \lambda-\dfrac{2\lambda-1}{1-\lambda}A(x), & \text{when } \lambda\in[½, 1) \text{ and } A(x)\in[0, 1-\lambda) \quad \text{(b)} \\ 1-\dfrac{1-2\lambda}{\lambda}A(x)-\lambda, & \text{when } \lambda\in(0, ½] \text{ and } A(x)\in[0, \lambda) \quad \text{(c)} \\ 1-\dfrac{1-2\lambda}{\lambda}(A(x)+\lambda-1)-\lambda, & \text{when } \lambda\in(0, ½] \text{ and } A(x)\in(1-\lambda, 1] \quad \text{(d)} \\ A(x), & \text{other} \quad \text{(e)} \end{cases}$$

(*iii*). Mapping $f^{\neg}: \{A(x) \mid x \in U\} \to [0, 1]$ confirms a fuzzy set $A^{\neg}$ on $U$, $A^{\neg}(x) = f^{\neg}(A(x)) = max(A^{\text{╕}}(x), A^{\sim}(x))$. Call $A^{\neg}$ the contradictory negation set of $A$.

(2) If $A$ is a clear set, then $A(x) \in \{0, 1\}$, $A^{\text{╕}}(x) = 1 - A(x)$, $A^{\sim}(x) = ½$, $A^{\neg}(x) = max(A^{\text{╕}}(x), A^{\sim}(x))$.

The set on the domain $U$ determined above is called "Sets with contradictory negation, opposite negation and intermediary negation", for short SCOI.

## 4.2 Logic with contradictory negation, opposite negation and intermediary negation

Based on SCOI, we further propose "logic LCOI+PLCOI with contradictory negation, contrary negation, and intermediary negation". LCOI+PLCOI is a formal logical calculus system that extends the syntax and semantics of classical logic. LCOI is propositional logic, and PLCOI is predicate logic.

### 4.2.1 Definition of the logic LCOI+PLCOI

Symbols ¬, ╕, and ~ denote "contradictory negation", "opposite negation" and "intermediary negation" respectively. Symbols ∨, ∧ and → denote 'disjunction', 'conjunction' and 'implication', respectively. Symbol "$\vdash$" denote formal deduction. The definition of logic LCOI+PLCOI is as follows

**Definition 1**. Let $\Im$ be set of atomic proposition. $\forall A \in \Im$, A is called well-formed formula (or formula). If A, B are formulas, then ¬A, ╕A, ~A, A→B, A∨B and A∧B are formulas.

(I) The following formulas as axioms:

(a1) $A \to (B \to A)$
(a2) $(A \to (A \to B)) \to (A \to B)$
(a3) $(A \to B) \to ((B \to C) \to (A \to C))$
(a4) $(A \to \neg B) \to (B \to \neg A)$
(a5) $(A \to \text{╕} B) \to (B \to \text{╕} A)$
(a6) $\neg A \to (A \to B)$
(a7) $((A \to \neg A) \to B) \to ((A \to B) \to B)$
(a8) $A \to A \vee B$
(a9) $B \to A \vee B$
(a10) $A \wedge B \to A$
(a11) $A \wedge B \to B$
(a12) $\text{╕} A \to \neg A \wedge \neg \sim A$, $\neg A \wedge \neg \sim A \to \text{╕} A$
(a13) $\sim A \to \neg A \wedge \neg \text{╕} A$, $\neg A \wedge \neg \text{╕} A \to \sim A$

(II) The deduction rules:

[D1] $A_1, A_2, \ldots, A_n \vdash A_i$ $(1 \le i \le n)$
[D2] $A \to B, A \vdash B$

The logic calculus formal system determined above is called "propositional logic with contradictory negation, opposite negation and intermediary negation", for short LCOI.

On the basis of LCOI, adding predicates, individual words, quantifiers ∀ and ∃, as well as the following axioms and deduction rule, we can constitute a predicate logic PLCOI with contradictory negation, opposite negation and intermediary negation.

(I) Axioms:

(a14) $\forall x A(x) \to A(a)$
(a15) $A(a) \to \exists x A(x)$
(a16) $\forall x(A(x) \to B) \to \exists x(A(x) \to B)$
(a17) $\text{╕} \forall x A(x) \to \exists x \text{╕} A(x)$, $\exists x \text{╕} A(x) \to \text{╕} \forall x A(x)$
(a18) $\text{╕} \exists x A(x) \to \forall x \text{╕} A(x)$, $\forall x \text{╕} A(x) \to \text{╕} \exists x A(x)$

(II) Deduction rule:

[D3] If $\Sigma \vdash A(a)$ ($\Sigma$ is the set of formulas), where the individual constant $a$ does not appear in $\Sigma$, then $\Sigma \vdash \forall x A(x)$.

The logic calculus formal system determined above is called "predicate logic with contradictory negation, opposite negation and intermediary negation", for short PLCOI.

The propositional logic LCOI and predicate logic PLCOI are denoted as LCOI+PLCOI.

### 4.2.2 A continuous-valued semantics of LCOI+PLCOI

Regarding the semantics of the logic LCOI+PLCOI, we previously provided a three-valued semantics and proved the soundness theorem, completeness theorem and compactness theorem for LCOI+PLCOI under this semantics [12,13]. In order to make LCOI+PLCOI applicable in practice, we hereby propose continuous-valued semantics for LCOI+PLCOI with a truth domain of [0, 1].

Let $\Sigma$ be a set of formulas in LCOI+PLCOI, and let A be a formula in LCOI+PLCOI. Since LCOI+PLCOI is a formal logical system, defining the formal deduction $\Sigma \vdash A$ is provable in LCOI+PLCOI, just as it is in other formal logic.

**Definition 1**. The formal deduction $\Sigma \vdash A$ ($\Sigma$ can be empty set) is provable in LCOI+PLCOI, if there exists a finite sequence of formulas $E_1, E_2, \ldots, E_n$ such that $E_n = A$ and for each $E_n$ ($1 \le k \le n$), either $E_k$ is an axiom in LCOI+PLCOI or $E_k$ follows from $E_i$ and $E_j$ ($i < k, j < k$) using the deduction rule in LCOI+PLCOI, then $E_1, E_2, \ldots, E_n$ is called a "*proof*" of $\Sigma \vdash A$, n *length* of *proof*. $\Sigma \vdash A$ is denoted $\vdash A$ when $\Sigma$ is empty.

**Definition 2** (continuous-valued interpretation). Let $\Im$ be set of all formulas in LCOI+PLCOI, $\lambda \in (0, 1)$. $\forall A \in \Im$, mapping $\partial: \Im \to [0, 1]$ is called a $\lambda$-assignment of $\Im$, consists of the individual domain D and the following assignments for each constant symbol, function symbol and predicate symbol in A:

(1) for each constant symbol, assign an object in *D* to correspond to it;

(2) for each n-variant function symbol, assign a mapping from $D^n$ to *D to* correspond to it;

(3) for each n-variant predicate symbol, assign a mapping from $D^n$ to [0, 1] to correspond to it, and

[1] If A is an atomic formula, $\partial(A)$ takes only one value from [0, 1];

[2] $\partial(A)+\partial(╕A) = 1$;

[3]
$$\partial(\sim A) = \begin{cases} \lambda - \dfrac{2\lambda-1}{1-\lambda}(\partial(A)-\lambda), & \text{when } \lambda\in[½, 1) \text{ and } \partial(A)\in(\lambda, 1] \quad \text{(a)} \\ \lambda - \dfrac{2\lambda-1}{1-\lambda}\partial(A), & \text{when } \lambda\in[½, 1) \text{ and } \partial(A)\in[0, 1-\lambda) \quad \text{(b)} \\ 1-\dfrac{1-2\lambda}{\lambda}\partial(A)-\lambda, & \text{when } \lambda\in(0, ½] \text{ and } \partial(A)\in[0, \lambda) \quad \text{(c)} \\ 1-\dfrac{1-2\lambda}{\lambda}(\partial(A)+\lambda-1)-\lambda, & \text{when } \lambda\in(0, ½] \text{ and } \partial(A)\in(1-\lambda, 1] \quad \text{(d)} \\ \partial(A), & \text{other} \quad \text{(e)} \end{cases}$$

[4] $\partial(\neg A) = \max(\partial(╕A), \partial(\sim A))$;

[5] $\partial(A\to B) = \Re(\partial(A), \partial(B))$. $\Re: [0, 1]^2 \to [0, 1]$ is a binary function;

[6] $\partial(A\vee B) = \max(\partial(A), \partial(B))$; $\partial(A\wedge B) = \min(\partial(A), \partial(B))$;

[7] $\partial(\forall xP(x)) = \min_{x\in D}\{\partial(P(x))\}$; $\partial(\exists xP(x)) = \max_{x\in D}\{\partial(P(x))\}$.

In Definition 2, how to determine the truth value $\partial(\sim A)$ of the intermediary negation $\sim A$ of formula A in [0, 1] (i.e., the expression (a)-(e)), the parameter variable $\lambda$ ($\lambda\in(0, 1)$) is the key to the definition. The basic idea is as follows:

Since the truth values $\partial(A)$, $\partial(\neg A)$, $\partial(╕A)$, $\partial(\sim A)\in[0, 1]$, in order to determine their value range in [0, 1], we introduce a parameter variable $\lambda\in(0, 1)$. Consequently, when $\lambda \ge ½$, [0, 1] is divided into three sub-intervals: $[0, 1-\lambda)$, $[1-\lambda, \lambda]$, $(\lambda, 1]$. If $\partial(A)\in(\lambda, 1]$, then based on [2] in the definition, $\partial(╕A)\in[0, 1-\lambda)$. At this point, if

∂(~A)∈[1–λ, λ], since (λ, 1] and [1–λ, λ] are disjoint intervals, then according to the principle that points in pairwise disjoint intervals in real variable functions have a one-to-one correspondence, the values in (λ, 1] correspond one-to-one with those in[1–λ, λ], and thus the expression (a) can be obtained. If ∂(A)∈[0, 1–λ) and ∂(~A)∈[1–λ, λ], we can similarly obtain expression (b). When λ ≤ ½, [0, 1] is divided into three sub-intervals: [0, λ), [λ, 1–λ], (1–λ, 1]. In the same manner, we can establish expressions (c) and (d). For other scenarios, ∂(~A) = ∂(A), which is expression (e).

For cases (a)–(d) in Definition 2, we illustrate them intuitively with the following figure (Figure 7). The symbols "•" and "o" in the figure represent the close endpoint and the open endpoint of an interval, respectively.

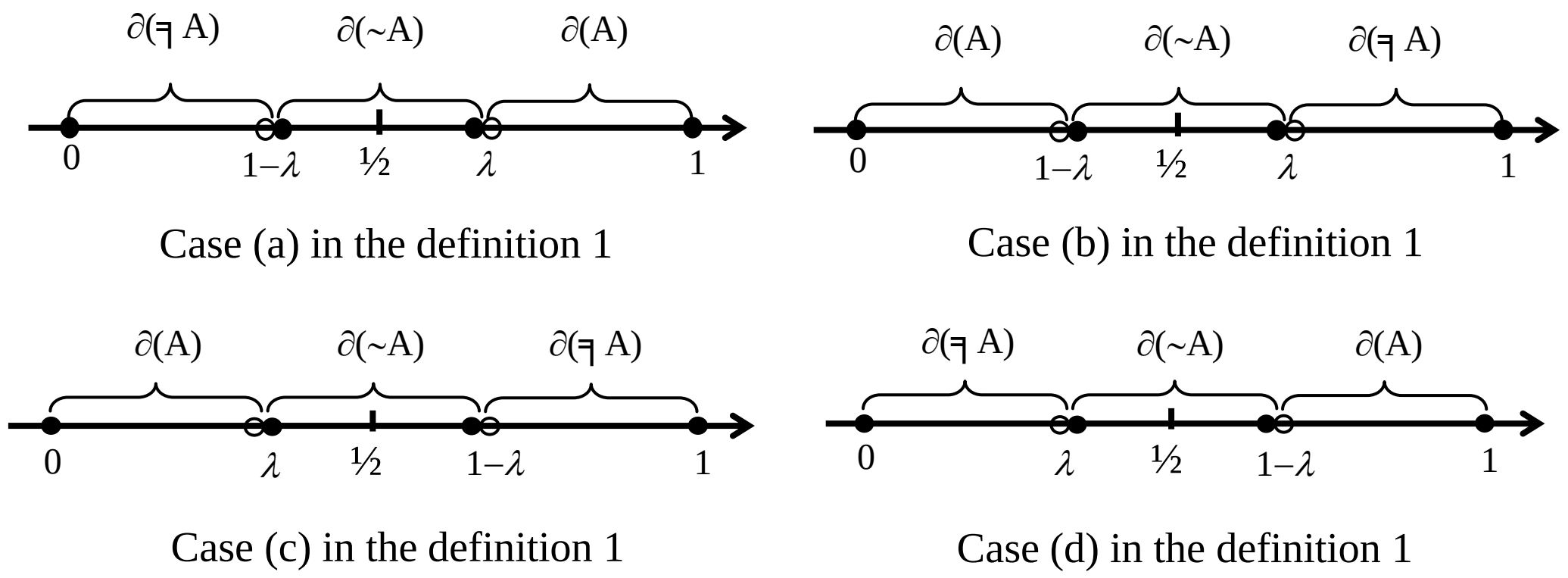

**Fig. 7** The interrelationships between ∂(A), ∂(╕A) and ∂(~A)

From the figure 7, it can be observed that:

(1) ∂(~A) serves as an "intermediary" between ∂(A) and its opposite ∂(╕A), reflecting the important philosophy idea that "all opposing concepts transition to each other through an intermediary between them" [28].

(2) λ∈(0, 1) is a variable parameter, it determines the value range of the membership degrees ∂(A), ∂(╕A) and ∂(~A). That is, λ is a "threshold" for the value range of these truth values. Its role and significance in practical applications will be discussed in detail in another article (a study on negation detection and negation resolution in medical texts).

As with the operational properties of SCOI [12, 13], it is easy to prove from Definition 2 that the truth values ∂(A), ∂(¬A), ∂(╕A) and ∂(~A) have the following relationships and properties.

**Proposition 1**. If ∂(A) = ½, then

$$\partial(\neg A) = \partial(╕A) = \partial(\sim A) = ½.$$

**Proposition 2**. If λ ≥ ½, then

$$\partial(A) > \partial(\sim A) > \partial(╕A) \text{ and } \partial(\neg A) = \partial(\sim A), \text{ if and only if } \partial(A)\in(\lambda, 1].$$

$$\partial(╕A) > \partial(\sim A) > \partial(A) \text{ and } \partial(\neg A) = \partial(╕A), \text{ if and only if } \partial(A)\in[0, 1-\lambda).$$

**Proposition 3**. If λ ≤ ½, then

$$\partial(A) > \partial(\sim A) > \partial(╕A) \text{ and } \partial(\neg A) = \partial(\sim A), \text{ if and only if } \partial(A)\in(1-\lambda, 1].$$

$$\partial(╕A) > \partial(\sim A) > \partial(A) \text{ and } \partial(\neg A) = \partial(╕A), \text{ if and only if } \partial(A)\in[0, \lambda).$$

**Proposition 4**.

$$\partial(\sim A)\in[1-\lambda, \lambda], \text{ when } \lambda \geq ½.$$

$$\partial(\sim A)\in[\lambda, 1-\lambda], \text{ when } \lambda \leq ½.$$

The above propositions conveys the following practical significance: for a formula (proposition) A in the logic system LCOI+PLCOI and its contradictory negation (¬A), opposite negation (╕A) and intermediary negation (~A), their truth values ∂(A), ∂(¬A), ∂(╕A) and ∂(~A) (values within [0, 1]) relate to each other under different threshold values of λ.

In the semantics of mathematical logic, the metatheorems (such as the soundness theorem and completeness theorem) revolve around tautologies. “Tautology” (or logical truth) is a core concept. In classical two-valued logic, the truth value of a formula is either 0 or 1. The formula A is a tautology if and only if its truth value is always 1. In non-classical logics like fuzzy logic, since truth values include multiple values such as 0 and 1, the concept of “tautology’ is weakened or modified (meaning A's truth value is not required to always be 1 but is required not to be lower than a certain value within the interval [0, 1]). This reflects different logical systems' interpretations of “logical truth”. Thus, for the logic LCOI+PLCOI, which studies crisp and fuzzy entities, we provide a definition of “$\lambda$-tautology” based on Definition 2.

**Definition 3** ($\lambda$-tautology). Let $\Gamma$ be set of $\lambda$-assignment of $\Im$, $\forall A \in \Im$. For any $\lambda$-assignment $\partial \in \Gamma$, if $\partial(A) = 1$ then A is called a tautology. If $\partial(A) \geq \lambda$ ($\lambda > ½$), A is called a $\lambda$-tautology, and denoted $\models$ A. If there exists a $\lambda$-assignment $\partial \in \Gamma$ such that $\partial(A) \geq \lambda$, then A is called $\lambda$-satisfiable.

As in the proof methods of mathematical logic, the axioms in a logical system must be proven to be tautologies in order to prove the metatheorems of the logical system, such as the soundness theorem and completeness theorem. Therefore, all axioms in LCOI+PLCOI should be $\lambda$-tautologies. To this end, we need to determine the binary function $\Re$ in the definition 2.

**Definition 4**. Let a, b$\in$[0, 1]. The mapping $\Re^o$: $[0, 1]^2 \rightarrow [0, 1]$ is $\Re$, if satisfies:

$\Re^o(a, b) = 1$, when $a \leq b$. (1)

$\Re^o(a, b) = \max(1-a, b)$, when $a > b$. (2)

It can be easily proven that $\Re^o$ has the following properties.

**Proposition 5**. Let a, b$\in$[0, 1]. Then

$a \geq b$, if and only if $\Re^o(a, c) \leq \Re^o(b, c)$， for all c$\in$[0, 1]。 (3)

$a > b$, if and only if $\Re^o(c, a) \geq \Re^o(c, b)$。 for all c$\in$[0, 1]。 (4)

**Lemma 1**. For LCOI+PLCOI, if A and A$\rightarrow$B are $\lambda$-tautologies, then B is a $\lambda$-tautology.

Proof: Let $\Re = \Re^o$, A and A$\rightarrow$B are $\lambda$-tautologies. Suppose B is not a $\lambda$-tautology. Then, according to Definition 2, there exists a $\lambda$-assignment $\beta \in \Gamma$, $\beta(B) < \lambda$. Since A and A$\rightarrow$B are $\lambda$-tautologies, so there are $\beta(A) \geq \lambda$ and $\beta(A \rightarrow B) \geq \lambda$. By [6] in the definition, $\beta(A \rightarrow B) = \Re^o(\beta(A), \beta(B)) \geq \lambda$. Due to $\beta(A) \geq \lambda$ and $\beta(B) < \lambda$，所以 $\beta(A) > \beta(B)$. Because $\lambda > ½$, so $\Re^o(\beta(A), \beta(B)) = \max(1-\beta(A), \beta(B)) < \lambda$ by (2). That is, it contradicts $\Re^o(\beta(A), \beta(B)) \geq \lambda$. Therefore, B is a $\lambda$-tautology. □

Based on Definition 2, let a, b and c represent $\partial(A)$, $\partial(B)$ and $\partial(C)$ respectively. If $\Re = \Re^o$, we can prove the following conclusion.

**Lemma 2**. Each axiom in LCOI+PLCOI is a $\lambda$-tautology.

Proof: Let $\Re = \Re^o$. According to [6] and Definition 3, the axioms (a1), (a2), and (a3) in LCOI+PLCOI can be expressed as follows:

(a1): $\Re(a, \Re(b, a)) \geq \lambda$, ($\lambda > ½$)

(a2): $\Re(\Re(a, \Re(a, b)), \Re(a, b)) \geq \lambda$, ($\lambda > ½$)

(a3): $\Re(\Re(a, b), \Re(\Re(b, c), \Re(a, c))) \geq \lambda$, ($\lambda > ½$)

For (a1). (i) If $a \leq b$, then $\Re(a, \Re(b, a)) = \Re(a, \max(1-b, a))$ by the definition 4, where if $1-b > a$, then $\Re(a, \max(1-b, a)) = \Re(a, 1-b) = 1 \geq \lambda$; if $1-b \leq a$, then $\Re(a, \max(1-b, a)) = \Re(a, a) = 1 \geq \lambda$. (ii) If $a > b$, then $\Re(b, a) \geq \Re(b, b)$ according to (4), $\Re(b, a) = 1$ by (1). So $\Re(a, \Re(b, a)) = \Re(a, 1)$. $\Re(a, \Re(b, a)) = 1$ by (1), i.e. $\Re(a, \Re(b, a)) \geq \lambda$. Therefore, the axiom (a1) is a $\lambda$-tautology by (i) and (ii).

For (a2). (i) If $a \leq b$, then $\Re(a, \Re(a, b)) \geq \Re(b, \Re(a, b))$ by (3). According to (1), $\Re(a, b) = 1$. Hence, $\Re(\Re(a, \Re(a, b)), \Re(a, b)) = \Re(\Re(a, 1), 1)$. $\Re(\Re(a, 1), 1) = \Re(1, 1) = 1$ by (1), i.e. $\Re(\Re(a, \Re(a, b)), \Re(a, b)) \geq \lambda$. (ii) If $a > b$, based on the above proof, it is only necessary to prove $\Re(a, \Re(a, b)) \leq \Re(a, b)$.

Suppose $\Re(a, \Re(a, b)) > \Re(a, b)$. $\Re(a, b) > b$ according to (4). $\Re(a, b) = \max(1-a, b) > b$ by (2). Hence, $\Re(a, b) = 1-a$. Substituting the hypothesis, there is $\Re(a, 1-a) > 1-a$, where if $a \leq 1-a$, then $\Re(a, 1-a) = 1$ by (1); if $a > 1-a$, then $\Re(a, 1-a) = \max(1-a, 1-a) = 1-a$. $\Re(a, 1-a) = 1-a$ contradicts $\Re(a, 1-a) > 1-a$. Thus, $\Re(a, \Re(a, b)) \leq \Re(a, b)$ holds true.

Therefore, the axiom (a2) is a $\lambda$-tautology by (i) and (ii).

For (a3). According to (1), only necessary to prove $\Re(a, b) \leq \Re(\Re(b, c), \Re(a, c))$. Suppose $\Re(a, b) > \Re(\Re(b, c), \Re(a, c))$. From this, $\Re(\Re(b, c), \Re(a, c)) \neq 1$. According to (1), so

$$\Re(b, c) > \Re(a, c) \qquad (5)$$

Thus, $\Re(a, c) \neq 1$. $a > b$ by (3), $a > c$ by (1). According to (2), $\Re(a, c) = \max(1–a, c)$. Substituting (5), $\Re(b, c) > \max(1–a, c)$, i.e. $b \leq c$ or $b > c$, there is $\Re(b, c) > \max(1–a, c)$.

If $b \leq c$, $\Re(b, c) = 1$ according to (1). Hence, $\Re(b, c) > \max(1–a, c)$.

If $b > c$, $\Re(b, c) = \max(1–b, c)$ according to (2). Because of $a > b$, i.e. $1–a < 1–b$, Hence, $\Re(b, c) = \max(1–b, c) > \max(1–a, c)$. However, if $1–b \leq c$, then $\max(1–b, c) = c$ and $\max(1–a, c) = c$, it contradicts $\max(1–b, c) > \max(1–a, c)$. So, only when $b > c$ and $1–b > c$, $\Re(b, c) > \max(1–a, c)$.

Substitute $a > b$, $b > c$, $1–b > c$ into the suppose: $\Re(a, b) > \Re(\Re(b, c), \Re(a, c))$, then $\max(1–a, b) > \max(1–\max(1–b, c), \max(1–a, c)) = \max(b, \max(1–a, c))$ by (3). However, when $1–a < b$, there is $\max(1–a, b) = b > \max(b, \max(1–a, c)) = b$, the two are contradictory; when $1–a \geq b$, there is $\max(1–a, b) = 1–a > \max(b, \max(1–a, c)) = 1–a$, the two are contradictory. Thus, the assumption $\Re(a, b) > \Re(\Re(b, c), \Re(a, c))$ is not valid. Therefore, the axiom (a3) is a $\lambda$-tautology.

Similarly, it can be proven that if $\Re = \Re^{o}$, then the axioms (a4) – (a18) in LCOI+PLCOI are all $\lambda$-tautologies. □

Based on the above results, similar to the proof methods (method of induction) for soundness in mathematical logic, we can prove the following soundness theorem for LCOI+PLCOI.

**Theorem 1** (*Soundness theorem*). Let $\Phi$ ($\Phi \subseteq \Im$) be a set of formulas in LCOI+PLCOI and A be a formula in LCOI+PLCOI.

(a) If $\vdash$ A, then $\models$ A.

(b) If $\Phi \vdash$ A, then $\Phi \models$ A.

Proof: If $\vdash$ A, then A is provable in LCOI+PLCOI. According to the definition 1, Induct on the length *n* of the sequence of formulas $A_1, A_2, \ldots, A_n$ for the proof of A.

(i) When $n = 1$, then according to Definition 1 in Section 5.3, $A_1$ (which is A) is an axiom in LCOI+PLCOI. By Lemma 2, A is a $\lambda$-tautology.

(ii) Suppose the theorem holds for $k < n$. That is, all proof sequences of A with fewer than n steps are $\lambda$-tautologies. We prove that the theorem holds when $n = k$. According to Definition 1 in Section 5.3, there are two cases: (1) $A_n$ is an axiom of LCOI+PLCOI, or (2) $A_n$ is a formula deduced from $A_i$ and $A_i$ and $A_j$ ($i < n, j < n$) using the deduction rule [D2] in LCOI+PLCOI. If it is case (1), then it is similar to the proof of (i). If it is case (2), the formal expressions of the formulas $A_i$ and $A_j$ must be B and B→A. According to the hypothesis, B and B→A are $\lambda$-tautologies. Therefore, by Lemma 1, A (i.e. $A_n$) is a $\lambda$-tautology. According to the principle of mathematical induction and Definition 3, $\models$ A holds.

When $\Phi$ is the empty set, (b) is equivalent to (a). Therefore, (b) can be proven similarly. □

We need to point out that when $\Re = \Re^{o}$, it can be verified that the completeness theorem for LCOI+PLCOI does not hold. Whether there exists a specific $\Re$ such that the completeness theorem for LCOI+PLCOI holds will be discussed in another article.

## 5. Triple with three types of negation

In Section 3, we conceptually propose that there are three different types of negation within triples and their elements: contradictory negation, opposite negation, and intermediary negation. Therefore, it is necessary to extend the concept of triples based on a logical system that can distinguish and express different negations both syntactically and semantically.

In this section, we distinguish triples into two types and discuss the basic characteristics of their negations.

Based on the set SCOI and the logic LCOI+PLCOI, we extend the triple and propose a triple that can fully express the three different types of negation of triple and their elements: “TCOI triple with contradictory negation, opposite negation and intermediary negation”.

## 5.1 Two Types of Triples and Their Negation Characteristics

For the triple <s, p, o>, from the perspective of structure and relations, it expresses the relationship established between the subject s and the object o through the predicate p. A triple represents a statement or an assertion, and the semantics of the statement is determined by the meanings of its elements. Since statements described in natural language can be distinguished into clear statements and fuzzy statements, triples should be classified into two categories: clear triple and fuzzy triple. Clear triple represent true/false statements, with the truth value domain of the statements being {0, 1}. Fuzzy triple represent statements that can be partially true or false, with the truth value domain being [0, 1].

*Clear triple*: In a triple <*s*, *p*, *o*>, if the elements *s*, *p* and *o* are clearly expressed (with clear meanings) and have no ambiguity within a specific domain, the triple is called a clear triple. A clear triple expresses a clear statement. For example, the triples <USA, president, Trump> and <constant x, nature, positive integer> are considered clear triples because all elements in these triples are clear concepts. They express clear statements: “the president of the USA is Trump” and “the constant x is a positive integer”.

*Fuzzy triple*: In a triple <*s*, *p*, *o*>, if one or more elements withen *s*, *p* and *o* are ambiguously expressed (with unclear meanings), the triple is called a fuzzy triple. A fuzzy triple expresses a fuzzy statement. For example, <tall person, likes, basketball> (here, the subject ‘tall person’ is a fuzzy concept), <Beijing, temperature, cold> (here, the object ‘cold’ is a fuzzy concept), and <Xiaoming, good at, programming> (here, the predicate ‘good at’ is a fuzzy concept). These express fuzzy statements: ‘tall people like basketball”, “the temperature in Beijing is cold” and “Xiaoming is good at programming”.

In short, for a triple, if all elements in the triple are clearly expressed, then the triple is a clear triple. If any element is expressed fuzzily, then the triple is a fuzzy triple.

Negation in triple and its application in knowledge representation and reasoning hold significant importance. It can be said that negation in triple is the indispensable "other half" in knowledge representation. It enables systems (e.g., Resource Description Frameworks, knowledge graphs) to extend from only describing what exists to also describing what does not exist, thus making logic more complete, knowledge more precise, and reasoning deeper. The importance of negation for triple is mainly reflected in the following aspects: (1) Enhancing data expressiveness and information integrity: Negation allows us to express the opposite of facts. (2) Supporting complex reasoning: Negation helps knowledge systems identify and handle counterexamples, enabling effective derivation. (3) Handling contradictions and consistency checks: Negation aids in identifying potential contradictions in information within the knowledge system. Through negation, conflicts in the system's knowledge can be detected and resolved, ensuring the accuracy of the knowledge system. (4) Expressing uncertainty and complex scenarios: Negation assists the system in handling more finely complex conditions and diverse data, such as expressing fuzzy queries or commands through negation.

From the perspective of the logical foundations of triple, classical logic and fuzzy logic form the basis for the formal syntax and fundamental deductive rules of clear (fuzzy) triple and their negations. Essentially, clear (fuzzy) triple correspond respectively to atomic propositions in classical predicate logic and fuzzy predicate logic. Since the formal languages of classical (fuzzy) predicate logic include only one kind of negation (classical negation), the negations within triple and their elements can only be expressed using classical negation in terms of logical syntax. The syntax and semantics of classical (fuzzy) first-order predicate logic determine this logical characteristic of clear (fuzzy) triple.

From the perspective of the semantic level of the triple structure, the negativity of clear and fuzzy triple exists in two forms: negation at the element level and negation at the triple level. In practice, these two forms of negation are not limited to a single type but involve three distinct types of negation with different meanings.

**Example 1**. In triple expressions, for the clear triple <*x*, property, positive number> (i.e., the statement "*x* is a positive number"), the triples < *x*, property, non-positive number > ("*x* is not a positive number"), < *x*, property, negative number> ("*x* is a negative number") and <*x*, property, zero> ("*x* is zero") represent its three different types of negation. For the fuzzy triple <Beijing, temperature, cold> (i.e., the statement "the temperature in Beijing is cold"), the triples <Beijing, temperature, not cold>, <Beijing, temperature, hot>, and <Beijing, temperature, warm> represent its three different types of negation.

**Example 2**. In element expression, for the subject "red" in the clear triple <red, is, color>, with the color relationships of the artistic color wheel (RYB) as the background, 'non-red', 'green' and 'yellow' represent three different types of negation for the subject "red". For the object "handsome" in the fuzzy triple <Chaplin, appearance, handsome>, 'not handsome', 'ugly' and 'ordinary' represent three different types of negation for the object.

For the three distinct forms of negation present in clear and fuzzy triples and their elements, classical logic and fuzzy logic, which only have one form of negation (classical negation) in formal language, cannot distinguish and express them. Therefore, to accurately and precisely express these different negations in knowledge representation, the classical triple <s, p, o> needs to be extended based on a non-classical or extended logic that can distinguish and express different negations at both the syntactic and semantic interpretation levels. Such logic includes multiple negation operators or semantic clarification mechanisms to achieve complete expression of negation meanings and accurate semantic modeling.

## 5.2 Triple with contradictory negation, opposite negation and intermediary negation

The intuitive meaning of a triple is the relationship established between the subject and the object as defined by the property, and the uncertainty within a triple is inevitably reflected by the uncertainty that exists in the relationship between subject and object [29]. Therefore, establishing the relationship between the subject and object is key in the clear and fuzzy triples.

For the clear triple and fuzzy triple the relationship between the subject and the object is distinguished as follows:

- If both the subject in the subject domain and the object in the object domain are clearly expressed (with clear meanings), then the relationship between the subject and the object is a binary clear relation: a composite mapping from the subject domain to the object domain, and then to {0, 1}.
- If either the subject in the subject domain or the object in the object domain, or both, have fuzzy expressions (with fuzzy meanings), then the relationship between the subject and the object is a binary fuzzy relation: a composite mapping from the subject domain to the object domain, and then to [0, 1].

The above two binary relations can be mathematically expressed as follows.

Let $X$ be a subject domain and $Y$ be an object domain.

(1) For $x \in X$, if exists $y \in Y$ and $y$ is clear expression, then relationship between $x$ and $y$ is a composite mapping $\mu$ from $X$ to $Y$ and $Y$ to $\{0, 1\}$, $\mu = f_2 \circ f_1$. $f_1 : X \to Y$ ; $f_2 : Y \to \{0, 1\}$. $\mu(x) = f_2(y) = f_2(f_1(x)) \in \{0, 1\}$.

(2) For $x \in X$, if exists $y \in Y$ and $y$ is fuzzy expression, then relationship between $x$ and $y$ is a composite mapping $\mu$ from $X$ to $Y$ and $Y$ to $[0, 1]$, $\mu = f_2 \circ f_1$. $f_1 : X \to Y$ ; $f_2 : Y \to [0, 1]$. $\mu(x) = f_2(y) = f_2(f_1(x)) \in [0, 1]$.

It can be seen that $\mu$ expresses the relationship between a subject $x$ and object $y$, $\mu(x)$ reflects the relatedness between $x$ and $y$, which we refer to as the "correlation degree" between $x$ and $y$.

For $\mu(x)$, we let the mapping $f$ in the definition of the set SCOI (Definition 1 in Section 4.1) be $\mu$. Thus, in SCOI, the membership degree $y(x)$ of $x$ to $y$ is the correlation degree $\mu(x)$ between $x$ and $y$. In other words, the correlation degree $\mu(x)$ between a subject $x$ and an object $y$ can be obtained by solving the membership degree $y(x)$ in the set SCOI.

According to the above, based on the set SCOI and the logic LCOI+PLCOI, we propose a triple that can fully express the three different types of negation of triple and their elements: “Triple with contradictory negation, opposite negation and intermediary negation”.

**Definition 1** (TCOI triple). Let $S$ be a subject domain, $O$ be an object domain, and $P$ be set of attributes. For any $s \in S$, $p \in P$ and $o \in O$, triple:

$$<s*, p*, (o*, \mu_{o*}(s*))>$$

is called the triple with contradictory negation, opposite negation and intermediary negation, for short TCOI triple. It expresses a statement (assertion): “$s$* and $o$* have a relation expressed by $p$* with a correlation degree $\mu_{o*}(s*)$”. Where, $s* \in \{s, \neg s, ╕ s, \sim s\} \subseteq S$, $p* \in \{p, \neg p, ╕ p, \sim p\} \subseteq P$, $o* \in \{o, \neg o, ╕ o, \sim o\} \subseteq O$. $\neg s$, $\neg p$ and $\neg o$ respectively are contradictory negation of $s$, $p$ and $o$; $╕ s$, $╕ p$ and $╕ o$ respectively are opposite negation of $s$, $p$ and $o$; $\sim s$, $\sim p$ and $\sim o$ respectively are opposite negation of $s$, $p$ and $o$. $\mu_{o*}(s*)$ is correlation degree betweens $s*$ to $o*$.

In the TCOI triple, if $s*$, $p*$ and $o*$ are all clearly expressed, then TCOI triple is a clear triple and $\mu_{o*}(s*) \in \{0, 1\}$. If there exists a fuzzy expression in $s*$, $p*$ and $o*$, then TCOI triple is a fuzzy triple and $\mu_{o*}(s*) \in [0, 1]$.

The TCOI triple is defined based on the set SCOI and the logic LCOI+PLCOI with three kinds of negation. It is capable of expressing three different negations of itself and its elements. Within the framework of the set SCOI and the logic LCOI+PLCOI, for the three negations of the TCOI triple and their meanings, as well as the triple representations for the three negations of elements and their meanings, we discuss as follows:

(i) For the three negations of the TCOI triple, their representations and meanings are as follows:

$\neg <s*, p*, (o*, \mu_{o*}(s*))>$: the contradictory negation of $<s*, p*, (o*, \mu_{o*}(s*))>$

$╕ <s*, p*, (o*, \mu_{o*}(s*))>$: the opposite negation of $<s*, p*, (o*, \mu_{o*}(s*))>$

$\sim <s*, p*, (o*, \mu_{o*}(s*))>$: the intermediary negation of $<s*, p*, (o*, \mu_{o*}(s*))>$

(ii) For the subject $s$ and its three negations $\neg s$, $╕ s$ and $\sim s$, the TCOI triple representation and its meaning are as follows:

$<s, p*, (o*, \mu_{o*}(s))>$:

$s$ and $o$* have a relation expressed by $p$* with a correlation degree $\mu_{o*}(s)$.

$<\neg s, p*, (o*, \mu_{o*}(\neg s))>$:

the contradictory negation $\neg s$ of $s$ and $o$* have a relation expressed by $p$* with a correlation degree $\mu_{o*}(\neg s)$.

$<╕ s, p*, (o*, \mu_{o*}(╕ s))>$

the opposite negation $╕ s$ of $s$ and $o$* have a relation expressed by $p$* with a correlation degree $\mu_{o*}(╕ s)$.

$<\sim s, p*, (o*, \mu_{o*}(\sim s))>$

the intermediary negation $\sim s$ of $s$ and $o$* have a relation expressed by $p$* with a correlation degree $\mu_{o*}(\sim s)$.

(iii) For the predicate $p$ and its three negations $\neg p$, $╕ p$ and $\sim p$, the TCOI triple representation and its meaning are as follows:

$<s*, p, (o*, \mu_{o*}(s*))>$;

$s$* and $o$* have a relation expressed by $p$ with a correlation degree $\mu_{o*}(s*)$.

$<s*, \neg p, (o*, \mu_{o*}(s*))>$

$s$* and $o$* have a relation expressed by the contradictory negation $\neg p$ of $p$ with a correlation degree $\mu_{o*}(s*)$.

$<s*, ╕ p, (o*, \mu_{o*}(s*))>$

$s$* and $o$* have a relation expressed by the opposite negation $╕ p$ of $p$ with a correlation degree $\mu_{o*}(s*)$.

$<s*, \sim p, (o*, \mu_{o*}(s*))>$:

$s$* and $o$* have a relation expressed by the intermediary negation $\sim p$ of $p$ with a correlation degree $\mu_{o*}(s*)$.

(iv) For the object $o$ and its three negations $\neg o$, $╕ o$ and $\sim o$, the TCOI triple representation and its meaning are as follows:

$<s*, p*, (o, \mu_o(s*))>$:

$s*$ and $o$ have a relation expressed by $p*$ with a correlation degree $\mu_o(s*)$.

$<s*, p*, (\neg o, \mu_{\neg o}(s*))>$:

$s*$ and the contradictory negation $\neg o$ of $o$ have a relation expressed by $p*$ with a correlation degree $\mu_{\neg o}(s*)$.

$<s*, p*, (\urcorner o, \mu_{\urcorner o}(s*))>$:

$s*$ and the opposite negation $\neg o$ of $o$ have a relation expressed by $p*$ with a correlation degree $\mu_{\urcorner o}(s*)$.

$<s*, p*, (\sim o, \mu_{\sim o}(s*))>$:

$s*$ and the intermediary negation $\neg o$ of $o$ have a relation expressed by $p*$ with a correlation degree $\mu_{\sim o}(s*)$.

We should point out that according to the semantics of the TCOI triple, the meanings of the three negations of a TCOI triple correspond respectively to the meanings of the triples with the three negations of its object *o*. That is, (i) is the same as the corresponding triplet in (iv).

(a) The contradictory negation of the TCOI triple is the same as the triple of the contradictory negation of its object *o*.

$$\neg <s*, p*, (o*, \mu_{o*}(s*))> = <s*, p*, (\neg o, \mu_{\neg o}(s*))>$$

(b) The opposite negation of the TCOI triple is the same as the triple of the opposite negation of its object *o*.

$$\urcorner <s*, p*, (o*, \mu_{o*}(s*))> = <s*, p*, (\urcorner o, \mu_{\urcorner o}(s*))>$$

(c) The intermediary negation of the TCOI triple is the same as the triple of the intermediary negation of its object *o*.

$$\sim <s*, p*, (o*, \mu_{o*}(s*))> = <s*, p*, (\sim o, \mu_{\sim o}(s*))>$$

For example, for a TCOI triple: <integer *x*, property, (positive integer, $\mu_o(x)$)> (the statement "integer x is a positive integer"), its contradictory negation ($\neg$), opposite negation ($\urcorner$) and intermediary negation (~) are represented respectively as follows:

(1) $\neg$ <integer *x*, property, (positive integer, $\mu_o(x)$)>

(2) $\urcorner$ <integer *x*, property, (positive integer, $\mu_o(x)$)>

(3) ~ <integer *x*, property, (positive integer, $\mu_o(x)$)>

In this TCOI triple, for the object "positive integer", the terms 'non-positive integer', 'negative integer' and 'zero' are its contradictory negation, opposite negation and intermediary negation, respectively. The triples with these three negations as the object and the statements they express are as follows:

(4) <*x*, property, (non positive integer, $\mu_{\neg o}(x)$)>: "integer *x* is not a positive integer"

(5) <*x*, property, (negative integer, $\mu_{\urcorner o}(x)$)>: "integer *x* is a negative integer"

(6) <*x*, property, (zero, $\mu_{\sim o}(x)$)>: "integer *x* is zero"

It can be seen that among the above six triples, the meanings of (1) and (4), (2) and (5), and (3) and (6) are the same.

For the correlation degree $\mu_{o*}(s*)$ in the TCOI triple, it can be determined either by the λ-assignment $\partial$ of the continuous value semantics in the logic LCOI+PLCOI (Definition 2 in Section 4.2.2), or by the membership degree in the definition of the set SCOI (Definition 1 in Section 4.1). The correlation degree $\mu_{o*}(s*)$ reflects the extent to which there is a relationship between the subject and the object.

**Example 3**. For the clear statement "the president of the United States is Trump", its TCOI triple can be expressed as <U.S, president, (Trump, 1)>. Here $\mu_o(s) = 1$, indicating that Trump's relevance to the president of U.S is at its highest. For the fuzzy statement "Churchill is somewhat overweight", its TCOI triple can be expressed as <Churchill, body type, (overweight, 0.8)>. Here $\mu_o(s) = 0.8$, indicating that Churchill's relevance to being overweight is 0.8, meaning that the degree of Churchill's overweight is 0.8.

From the above, it is evident that the syntactic form and semantics of the TCOI triple differ from those of a general triple. TCOI not only distinguishes between crisp triples and fuzzy triples as well as the relationships and degrees of association between the subject and the object, but also is capable of expressing contradictory negation, opposite negation, and intermediary negation in the triple and its elements.

TCOI triple is a semantic and structural extension of the classical triple <s, p, o>. While retaining the ability to express affirmative assertions, it systematically introduces the three semantic dimensions of contradictory negation, opposite negation and intermediary negation, allowing these negations to independently apply to the elements (s, p, o) of the triple and on the whole triple. This significantly enhances the triple model capability to represent and reasoning about complex negative information.

We believe that, for the classical triple, the TCOI triple extends it in three dimensions: semantic extensibility, structural extensibility, and logical positioning. It has the following characteristics:

- The TCOI triple not only retains the expressive capability of the classical triple but also additionally introduces three mutually independent and semantically heterogeneous negation dimensions. This constitutes a substantial semantic extension.
- The TCOI triple appends negation type labels onto the atomic structure of the classical triple, expanding it from an “unmarked affirmative structure” to a “multivariate structure with negation markings”. This is a crisp structural extension.
- The TCOI triple extends the classical triple, which is based on classical logic (with only one classical negation in its formal language), to a foundation based on the logic LCOI+PLCOI (whose formal language and semantics include contradictory negation, opposite negation and intermediary negation).

## 6. Expression ability of TCOI triple

“Negation” is an important and universal phenomenon in language, and a necessary feature of every human language [1]. In this section, we demonstrate the expressive power of the TCOI triple using examples of everyday sentences and Web resource description statements that involve different kinds of negation.

(I) Expression of different negations in everyday statements

**Example 1**. In everyday statements, the following types of sentences are common:

(1) *Clear statements*: “the integer *x* is neither a positive integer nor a negative integer”, “this profession is neither safe nor dangerous”, “this country belongs to neither the East nor the West”, and so on.

(2) *Fuzzy statements*: “S is neither tall nor short”, “my father is neither fat nor thin”, “there is neither more nor less salt in the dish”, and so on.

Obviously, this type of statement contains more complex negativity.

We take the statement “S is neither tall nor short” from (2) as an example, which is denoted as *Ex*.

*Ex* is a composite statement consisting of an atomic statement and its different negations. Among them, the atomic statement is “S is a tall person”. Based on the logic LCOI+PLCOI and the TCOI triple, this atomic statement and its different forms of negation expressions as well as the triple representations are as follows:

A: “S is a tall person” (i.e., an atom formula A in LCOI+PLCOI)

<S, height, (tall, $\mu_o$(S))>

$\neg$A: “S is not a tall person” (i.e., the contradictory negation $\neg$A of A)

<S, height, (not tall, $\mu_{\neg o}$(S))>

╕A: “S is a short person” (i.e., the opposite negation ╕A of A)

<S, height, (short, $\mu_{╕o}$(S))>

~A: “S is medium height” (i.e., the intermediary negation ~A of A)

<S, height, (medium, $\mu_{\sim o}$(S))>

$\neg$╕A: “S is not a short person” (i.e., the contradictory negation $\neg$╕A of ╕A)

<S, height, (not short, $\mu_{\neg ╕o}$(S))>

Therefore, according to the meaning of *Ex*, *Ex* is the conjunction of $\neg$A and $\neg$╕A. That is

$$Ex = \neg A \wedge \neg ╕ A$$

According to axiom (a13) in logic LCOI+PLCOI (Definition 1 in Section 4.2.1):

$$\sim A \to \neg A \wedge \neg ╕ A,\ \neg A \wedge \neg ╕ A \to \sim A$$

~A and ¬A ∧ ¬ ╕ A nd ¬A ∧ ¬ ╕ A mutually implication each other and are logically equivalent. That is,

$$\sim A \cong \neg A \wedge \neg ╕ A$$

This indicates that the intermediary negation ~A of A and *Ex* have the same meaning. That is, the statement "S is neither tall nor short" has the same meaning as the statement "S is medium height".

Regarding the statement *Ex* and the different negations of *Ex* and their relationships, we can illustrate them as follows (Fig. 8):

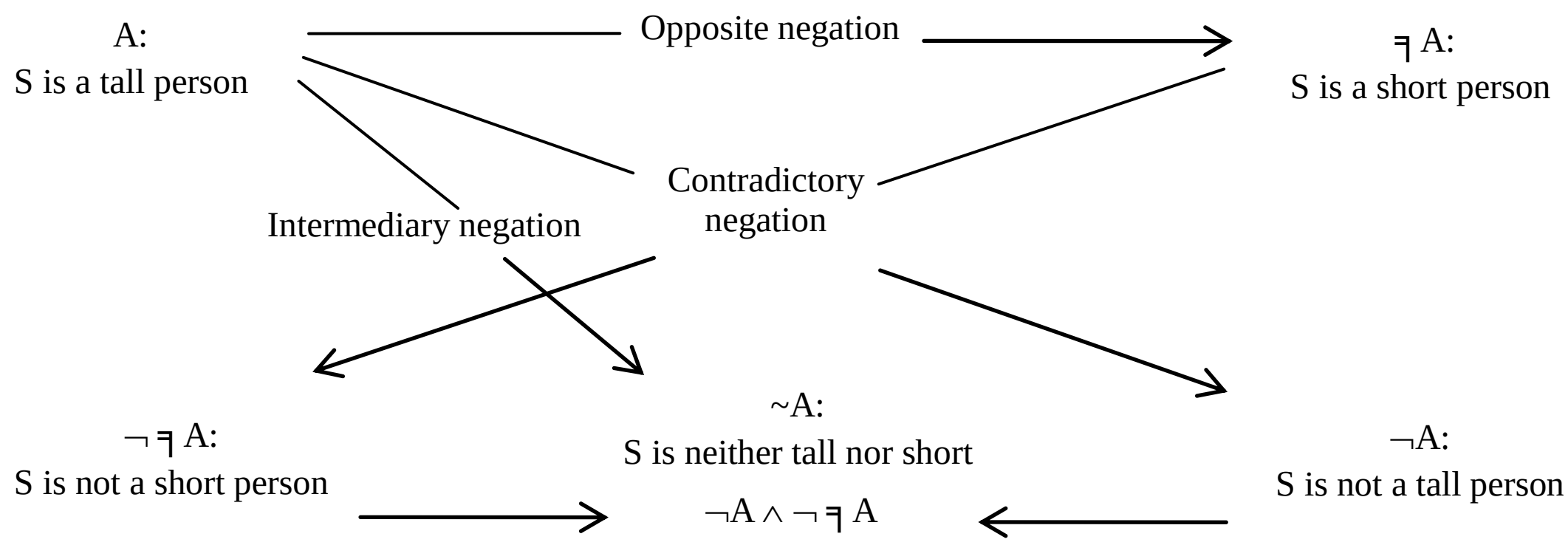

**Fig. 8** The statement *Ex* and the different negations within *Ex* and their relationships

Above, taking the fuzzy statement "S is neither tall nor short" in (2) as an example, we demonstrated that this statement can be expressed by the "intermediary negation" of the TCOI triple <S, height, (tall, $\mu_o$(S))>, namely <S, height, (medium, $\mu_{\sim o}$(S))>. For the clear statement in (1), similarly, based on the logic LCOI+PLCOI, expressing it with a clear TCOI triple leads to conclusions analogous to those above.

Thus, it can be seen that for the more complex negative statements of types (1) and (2), whether clear or fuzzy, they can all be expressed by an intermediary negation of a TCOI triple.

(II) Expression of different negations and fuzziness in Web resource description statements

**Example 2**. For the web resource: "Peking University Professor Li Song" (abbreviated as "Li"), use TCOI triple to describe his appearance.

Clearly, the attributes of a person include "date of birth", "home address", "marital status" and "appearance" etc. For the attribute of "appearance", its attribute values include 'beautiful', 'not beautiful', 'ugly' and 'ordinary'. It is evident that these attribute values represent different fuzzy concepts (fuzzy sets).

According to the set SCOI and the logic LCOI+PLCOI, the attribute values 'not beautiful', 'ugly' and 'ordinary' are respectively the contradictory negation, opposite negation and intermediary negation of 'beautiful'. Therefore, the appearance of the resource "Li" can be represented by the following four TCOI triple:

(1) < Li, appearance, (beautiful, $\mu_o$(Li))>
(2) < Li, appearance, (unbeautiful, $\mu_{\neg o}$(Li))>
(3) < Li, appearance, (ugly, $\mu_{╕ o}$(Li))>
(4) < Li, appearance, (ordinary, $\mu_{\sim o}$(Li))>

According to the definition of SCOI (Definition 1 in Section 4.1), if assuming $\lambda$ = 0.6, and the membership degree of "Li" to fuzzy set 'beautiful' is 0.9 (i.e., correlation degree $\mu_o$(Li) of resource "Li" to attribute value 'beautiful' is 0.9), it can be calculated from the definition of SCOI that the correlation degree of resource "Li" to attribute value 'ugly' is $\mu_{╕ o}$(Li) = 1– 0.9 = 0.1, the correlation degree of resource "Li" to attribute value 'ordinary' is $\mu_{\sim o}$(Li) = 0.45, and the correlation degree of resource "Li" to attribute value 'not beautiful' is $\mu_{\neg o}$(Li)

= 0.275. Therefore, the four different TCOI triple and their meanings are as follows:

<Li, appearance, (beautiful**,** 0.9)>: “Li’s appearance beautiful degree is 0.9”

<Li, appearance, (not beautiful, 0.275)>: “Li’ appearance not beautiful degree is 0.275”

<Li, appearance, (ugly, 0.1)>: “Li’s appearance ugly degree is 0.1”

<Li, appearance, (ordinary, 0.45)>: “Liu’s appearance ordinary degree is 0.45”

In the Resource Description Framework (RDF), an RDF graph is a set of RDF triples [1]. The RDF graph of four TCOI triple describing the appearance of “Li” (Fig. 9) is as follows:

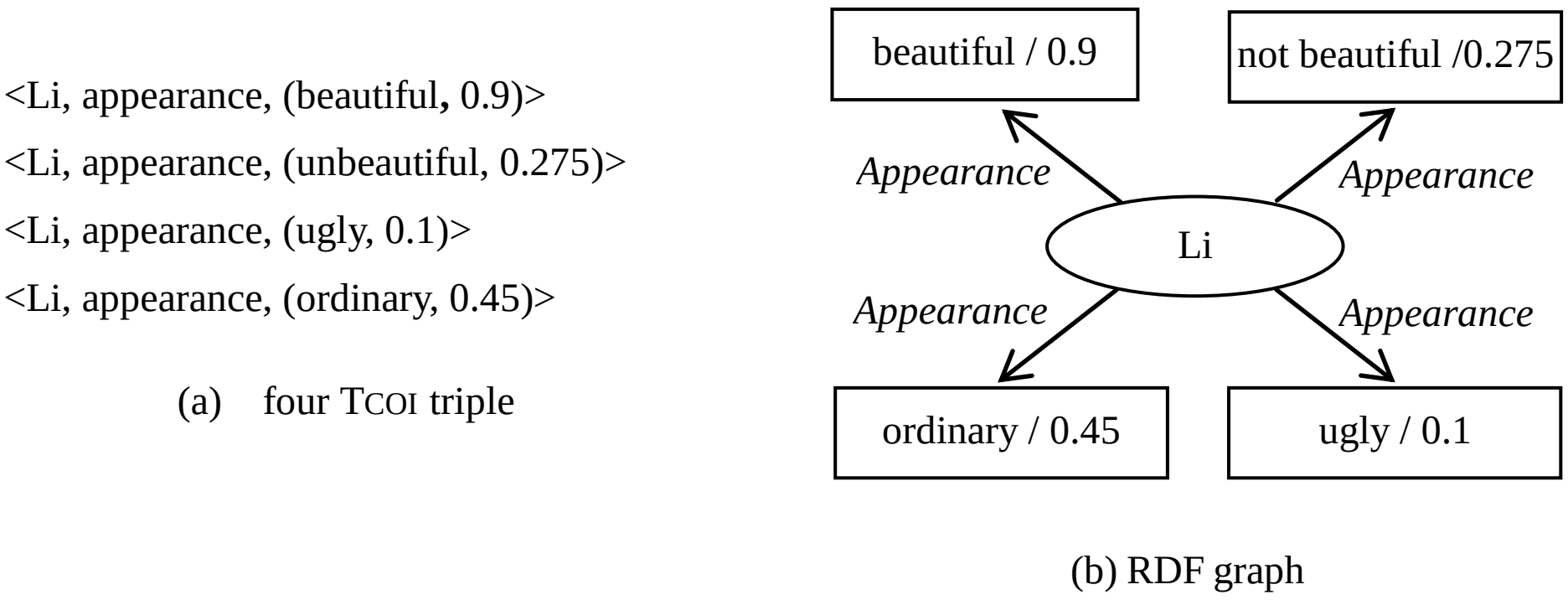

(b) RDF graph

**Fig. 9** TCOI triple and RDF graph

For the different negations and fuzziness in the aforementioned everyday statements and web resource description statements, if expressed using conventional triples, it would be more complex or difficult to convey them due to the inability to distinguish and handle the three distinct types of negation and their relationships present in those statements.

## 7. TCOI triple reasoning

A key feature of semantic data models is their support for reasoning. Triple reasoning is a semantic reasoning approach that relies on semantic understanding of data and infers new triples, i.e., new knowledge, from existing triples or sets of triples. Research on triple reasoning has evolved from symbolic logic to statistical learning and then to neural representation learning. Early studies on triple reasoning were mainly built upon logic programming, description logic, and rule-based database reasoning, where Horn rules, Datalog and description logic formed the basic formal frameworks for triple reasoning. With the development of the Semantic Web and RDF/OWL, researchers further focused on RDF semantic entailment, RDFS/OWL reasoning, and SPARQL recursive querying [30,31]. In recent years, knowledge graph embedding methods have transformed triple reasoning into a relation modeling problem in vector spaces, significantly improving the scalability of knowledge completion. Graph neural networks and differentiable rule learning methods further combine the advantages of symbolic reasoning and neural networks, becoming important research directions in current triple reasoning studies [32–36].

In triples reasoning, implication-based reasoning is a fundamental and typical semantic reasoning method. Under a given semantic interpretation, it derives new triple facts by exploiting the semantic entailment relations among existing triple or sets of triples. This approach is grounded in a clear logical foundation, with its core idea being to abstract observed entity–relation instances into first-order logical implications or Horn clauses, such as relation transitivity, symmetry, antisymmetry, and type constraints. On this basis, it constructs interpretable reasoning chains to infer latent triples from explicit facts. Implication-based reasoning exhibits strong symbolic interpretability and consistency, and it can explicitly characterize the logical dependencies among facts, thereby providing a more transparent basis for inference at the knowledge representation level. Overall, implication-based reasoning is not only a classic paradigm in triples reasoning, but also an important bridge between logical knowledge representation and data-driven learning, and thus it is widely used in the construction of reliable reasoning systems [37–41].

In this section, we mainly explore implication-based reasoning in TCOI triple reasoning. It discusses the relationship between TCOI triple implication reasoning and logic LCOI+PLCOI reasoning, as well as the logical foundation of TCOI triple implication reasoning. To demonstrate the capability of TCOI triple implication reasoning, we apply it to counterfactuals and counterfactual reasoning. Using an example of fuzzy counterfactual and counterfactual reasoning, it shows that the three types of fuzzy counterfactual reasoning based on different negations correspond to three kinds of fuzzy TCOI triple implication reasoning. Additionally, a truth-value (continuous value) algorithm is proposed for these implication reasonings, and calculations are performed on the reasoning example.

## 7.1 Logical foundation of TCOI triple implication reasoning

In triples reasoning, implication-based reasoning essentially means that if the premise (a set of triples) is true, then the conclusion (a triple or a set of triples) must also be true.

To support TCOI triple implication reasoning, based on the semantics of logic LCOI+PLCOI, we introduce the concept of "TCOI-entailment" as the "semantic entailment" in TCOI triple implication reasoning. Through TCOI -entailment, a connection is established between TCOI triple entailment reasoning and reasoning in logic LCOI+PLCOI. This indicates that the formally proven inference laws in logic LCOI+PLCOI are valid in TCOI triple entailment reasoning, and that LCOI+PLCOI provides the logical foundation for TCOI triple entailment reasoning.

**Definition 1** (TCOI-entailment). Let *T* be a TCOI triple or set of TCOI triple, *S* be a set of TCOI triple. *S* entails *T*, denoted as $S \models T$, if and only if every interpretation that satisfies *S* in the semantic interpretation of LCOI+PLCOI also satisfies *T*. If *S* and *T* entail each other, then *S* and *T* are logically equivalent. This entailment is called the semantic entailment in TCOI triple implication reasoning, abbreviated as TCOI-entailment.

Concise speaking, in TCOI triple implication reasoning, $S \models T$ means that *T* is a semantic consequence of *S* (i.e., *S* is the condition of semantic inference, and *T* is the conclusion).

A classic triple <s, p, o> is a minimal statement and is an atomic formula in classical predicate logic. Similarly, since the TCOI triple is defined based on the logic LCOI+PLCOI, a TCOI triple is also an atomic formula within the logic LCOI+PLCOI. Using the connectives (¬, ╕, ~, →, ∧, ∨, ∃, ∀) in LCOI+PLCOI, multiple TCOI triple can be combined into new TCOI triple, which correspond to compound formulas in the logic LCOI+PLCOI. Therefore, the *T* in Definition 1 is a formula or a set of formulas in the logic LCOI+PLCOI, and *S* is a set of formulas. Consequently, accordng to the soundness theorem of LCOI+PLCOI (Theorem 1 in Section 4.2.2), for the axioms, deduction rules and proven formal theorems in LCOI+PLCOI, the entailment $S \models T$ holds. Therefore, for TCOI triple implication reasoning, based on TCOI-entailment, we can derive the following conclusion:

(1) If *T* is an axiom in LCOI+PLCOI, then $\models T$.
(2) If *S* and *T* are respectively the premise and conclusion of a deduction rule in LCOI+PLCOI (i.e., $S \vdash T$), then $S \models T$.
(3) If *T* is conclusion of a proven formal theorem in LCOI+PLCOI (i.e., $\vdash T$), then $\models T$.
(4) If *S* and *T* are respectively the premise and conclusion of a proven formal theorem in LCOI+PLCOI (i.e., $S \vdash T$), then $S \models T$.

Therefore, the TCOI triple implication reasoning has the following properties:

**Property 1**. If *T* is axiom in the logic LCOI+PLCOI, then

⊨ A→(B→A); (A→(A→B))→(A→B); (A→B)→((B→C)→(A→C)); (A→¬B)→(B→¬A); (A→ ╕B)→(B→ ╕A); ¬A→(A→B); ((A→¬A)→B)→((A→B)→B); A→A∨B; B→A∨B; A∧B→A; A∧B→B; ╕A→ ¬A ∧

¬~A; ~A→ ¬A ∧¬ ╕A

╞ ∀xA(x)→A(a); A(a)→∃xA(x); ∃x(A(x)→B)→(A(a)→B); ╕∀xA(x)→∃x╕A(x); ∃x╕A(x)→ ╕∀xA(x); ╕∃xA(x)→∀x╕A(x); ∀x╕A(x)→╕∃xA(x)

(Note. remove "╞", which is the axioms in LCOI+PLCOI (the axioms (a1)-(a18) in Section 4.2.1)

**Property 2**. If *S* and *T* are respectively the premises and conclusions of the deduction rules in LCOI+PLCOI (i.e., *S*├ *T*), then

(1) $A_1, A_2, ..., A_n$ ╞ $A_i$ (i∈{1, 2, ..., n}).

(2) A→B, A╞ B.

(3) If $A_1, A_2, ..., A_n$ ╞ A(*a*), where *a* does not occur in $A_i$ (1≤ i ≤ n), then $A_1, A_2, ..., A_n$ ╞ ∀xA(x).

(Note. replace ╞ with ├, that is the deduction rules of LCOI+PLCOI (D1, D2 and D3 in Section 4.2.1)

**Property 3**. If *T* is conclusions of the proven formal theorems in LCOI+PLCOI (i.e., ├ *T*), then

[1] ╞ A→A; ((A→B)→C)→(B→C); A→((A→B)→(C→B)); A→((A→B)→B)

[2] ╞ (((A→B)→B)→C)→(A→C); (A→(B→C))→(B→(A→C)); (B→C)→((A→B)→(A→C))

[3] ╞ ((A→B)→(A→(A→C)))→((A→B)→(A→C))

[4] ╞ (B→(A→C))→((A→B)→(A→C)); (A→(B→C))→((A→B)→(A→C))

[5] ╞ ¬A→(A→ ¬(B→B)); B→((A→ ¬B)→¬A); (A→ ¬B)→((A→B)→ ¬A)

[6] ╞ (A→B)→((A→ ¬B)→ ¬A); (A→B)→(¬B→ ¬A); (A→ ¬A)→ ¬A; A→ ¬ ¬A

[7] ╞ (¬A→ ¬B)→(B→A); ¬ ¬A→(¬ ¬A→A); ¬ ¬A→A; (¬A→B)→((¬A→ ¬B)→A)

[8] ╞ ╕A→(A→ ╕(B→B)); B→((A→ ╕B)→ ╕A); (A→ ╕B)→(A→(B→ ╕(B→B)))

[9] ╞ (A→B)→((A→ ╕B)→ ╕A); (A→B)→(╕B→ ╕A); (A→ ╕A)→ ╕A; A→ ╕╕A

[10] ╞ (╕A→ ╕B)→(B→A); ╕╕A→(╕╕A→A); ╕╕A→A; ¬(A∧¬A); ¬(╕A∧~A)

[11] ╞ ¬(A∧~A); ¬(A∧╕A)

(Note. replace ╞ with ├, that is the formal theorems in LCOI+PLCOI (theorem 1- theorem 4, Section 4.2.1 in [13]))

**Property 4**. If *S* and *T* are respectively the premises and conclusions of the proven formal theorems in LCOI+PLCOI (i.e., *S*├ *T*), then

(1) A╞ A; ╕╕A; ¬¬A; ¬╕A∧¬~A; ¬╕~A

(2) ╕╕A╞ A;

(3) ¬ ¬A╞ A

(4) A, ╕B╞ ╕(A→B)

(5) ╕(A→B)╞ A, ╕B

(6) ~A╞ ~╕A

(7) ~╕A╞ ~A

(8) ¬╕A∧¬~A╞ A

(9) A, ¬A╞ B

(10) A, ╕A╞ B

(11) A, ~A╞ B

(Note. replace ╞ with ├, that is the formal theorems in LCOI+PLCOI (theorem 5 and theorem 6, Section 4.2.1 in [13])).

The above TCOI triple implication reasoning properties indicate that the proven formal inference relations in the logic LCOI+PLCOI are valid in the TCOI triple implication reasoning. This validity has the following significance:

- LCOI+PLCOI provide a solid logical foundation for the TCOI triple implication reasoning.
- For any formal deduce *S*├ *T* in logic LCOI+PLCOI, according to the soundness theorem of LCOI+PLCOI (Theorem 1 in Section 4.2.2), *S*╞ *T* holds in TCOI triple implication reasoning. That is, *S*├ *T* ⇒ *S*╞ *T*. Conversely, *S*╞ *T* ⇒ *S*├ *T* according to the completeness theorem of LCOI+PLCOI [8, 9].
- The TCOI triple has stronger reasoning capabilities than the classic triple and other extensions.

- This deepens the integration of logical theory and triple theory, and expands the scope of TCOI triple for data analysis and knowledge processing.

## 7.2 Applications of TCOI triple implication reasoning in counterfactual reasoning

Counterfactuals and counterfactual reasoning are cognitive processes in humans that constitute a significant topic of interest across several fields, including philosophy, cognitive science, linguistics, logic, and artificial intelligence. The core of counterfactual thinking lies in negating the facts that have occurred (the real world) to envision a possibility that has not happened (the possible world)[42]. A counterfactual refers to a hypothetical statement that is contrary to the facts, often used to describe a "what if things had not happened as they actually did" scenario. Counterfactual reasoning is a process that involves reasoning based on counterfactuals, with its basic structure being the negation of facts to construct a hypothetical premise that supports hypothetical reasoning. Negation occupies a central position in both counterfactual statements and reasoning, without negation, there would be no counterfactuals or counterfactual reasoning [43-48]. Crisp counterfactuals and fuzzy counterfactuals are two different types of conditional sentences, distinguished by significant differences in the premise or conclusion statements. The premise and conclusion of crisp counterfactuals are clear statements (i.e., the concepts contained are all clear concepts). The premise or the conclusion or both of fuzzy counterfactuals are fuzzy statements (i.e., they contain fuzzy concepts) [49].

Triple reasoning and counterfactual reasoning can both be classified as forms of semantics-based reasoning, but they focus on different levels of semantics. Triple reasoning primarily concerns relational semantics and structured semantics, whereas counterfactual reasoning mainly concerns causal semantics.

In this section, we apply TCOI triple implication reasoning to counterfactuals and counterfactual reasoning. Using an example of the counterfactuals and counterfactual reasoning based on three different kinds of negation, it show that three kinds of fuzzy counterfactual reasoning based on different negations correspond to three kinds of fuzzy TCOI triple implication reasoning.

**Example**. Fact F1: If the water temperature in the teacup is high, then the hand feels that the teacup is hot. There are following counterfactuals:

(1) If hand feels the teacup not hot, then the water temperature in the teacup is not high.

(2) If hand feels the teacup cool, then the water temperature in the teacup is low.

(3) If hand feels the teacup neither hot nor cool, then the water temperature in the teacup is neither high nor low.

It can be seen that "high", "not high", "low", "neither high nor low", "hot", "not hot", "cool" and "neither hot nor cool", are all fuzzy concepts. Therefore, the fact F1 and the counterfactuals (1), (2) and (3) are all fuzzy statements, that is (1), (2) and (3) are fuzzy counterfactuals.

According to the set SCOI and the logic LCOI+PLCOI, 'not high', 'low' and 'neither high nor low' are respectively the contradictory negation, opposite negation and intermediary negation of "high"; 'not hot', 'cool' and "neither hot nor cool" are respectively the contradictory negation, opposite negation and intermediary negation of "hot". Therefore, we may refer to conditionals (1), (2), and (3) as three fuzzy counterfactuals based on contradictory negation, opposite negation and intermediary negation, respectively.

For the statements in Fact F1 and fuzzy counterfactuals (1), (2), and (3), according to the definitions of the logic LCOI+PLCOI and TCOI triple, we can express them using TCOI triple as follows:

A: "the water temperature in the teacup is high", <teacup, temperature, (high, $\mu_o$(teacup))>

¬A: "the water temperature in the teacup is not high". <teacup, temperature, (not high, $\mu_{\neg o}$(teacup))>

╕A: "the water temperature in the teacup is low". <teacup, temperature, (low, $\mu_{╕o}$(teacup))>

~A: "the water temperature in the teacup is neither high nor low", <teacup, temperature, (neither high nor low, $\mu_{\sim o}$(teacup))>

B: “the hand feels the teacup hot”. <hand, feels, (hot, $\mu_o$(hand))>

¬B: “the hand feels the teacup not hot”. <hand, feels, (not hot, $\mu_{\neg o}$(hand))>

╕B: “the hand feels the teacup cool”. <hand, feels, (cool, $\mu_{╕o}$(hand))>

~B: “the hand feels the teacup neither hot nor cool”. <hand, feels, (neither hot nor cool, $\mu_{\sim o}$(hand))>

Therefore, the TCOI triple representations of fact F1 and fuzzy counterfactuals (1), (2) and (3) are as follows:

F1 <teacup, temperature, (high, $\mu_o$(teacup))> → <hand, feels, (hot, $\mu_o$(hand))>

(1) <hand, feels, (not hot, $\mu_{\neg o}$(hand))> → <teacup, temperature, (not high, $\mu_{\neg o}$(teacup))>

(2) <hand, feels, (cool, $\mu_{╕o}$(hand))> → <teacup, temperature, (low, $\mu_{╕o}$(teacup))>

(3) <hand, feels, (neither hot nor cool, $\mu_{\sim o}$(hand))>→ <teacup, temperature, (neither high nor low, $\mu_{\sim o}$(teacup))>

Counterfactual reasoning (*CR*) is a form of reasoning based on counterfactuals. Fuzzy counterfactual reasoning based on the three fuzzy counterfactuals (1), (2) and (3) is respectively as follows:

$CR_{\neg}$: If the water temperature in the teacup is high, then the hand feels the teacup is hot, but the hand feels the teacup is not hot. Therefore, the water temperature in the teacup is not high.

$CR_{╕}$: If the water temperature in the teacup is high, then the hand feels the teacup is hot, but the hand feels the teacup is cool. Therefore, the water temperature in the teacup is low.

$CR_{\sim}$: If the water temperature in the teacup is high, then the hand feels the teacup is hot, but the hand feels the teacup is neither hot nor cool. Therefore, the water temperature in the teacup is neither high nor low.

Based on the logic LCOI+PLCOI, the fuzzy counterfactual reasoning $CR_{\neg}$, $CR_{╕}$ and $CR_{\sim}$ can be formally represented as follows (the symbol ⇒ stands for logical deduction):

$CR_{\neg}$: A→B, ¬B ⇒ ¬A

$CR_{╕}$: A→B, ╕B ⇒ ╕A

$CR_{\sim}$: A→B, ~B ⇒ ~A

According to the logic LCOI+PLCOI and the definition of TCOI-entailment (Definition 1), these fuzzy counterfactual reasoning can be separately expressed using TCOI triple implication reasoning as follows:

$T_{\neg}$: A→B, ¬B ⊨ ¬A

<teacup, temperature, (high, $\mu_o$(teacup))> →<hand, feels, (hot, $\mu_o$(hand))>∧<hand, feels, (not hot, $\mu_{\neg o}$ (hand))>
⊨ <teacup, temperature, (not high, $\mu_{\neg o}$(teacup))>

$T_{╕}$: A→B, ╕B ⊨ ╕A

<teacup, temperature, (high, $\mu_o$(teacup))> → <hand, feels, (hot, $\mu_o$(hand))>∧<hand, feels, (cool, $\mu_{╕o}$(hand))>
⊨ <teacup, temperature, (low, $\mu_{╕o}$(teacup))>

$T_{\sim}$: A→B, ~B ⊨ ~A

<teacup, temperature, (high, $\mu_o$(teacup))> → <hand, feels, (hot, $\mu_o$(hand))> ∧<hand, feels, (neither hot nor cool, $\mu_{\sim o}$(hand))>
⊨ <teacup, temperature, (neither high nor low, $\mu_{\sim o}$(teacup))>

Here, $T_{\neg}$, $T_{╕}$ and $T_{\sim}$ respectively denote the three fuzzy TCOI triple implication reasoning based on contradictory negation ¬, opposite negation ╕ and intermediary negation ~.

It can be seen that fuzzy TCOI triple implication reasoning $T_{\neg}$, $T_{╕}$ and $T_{\sim}$, respectively correspond to the three fuzzy counterfactual reasoning $CR_{\neg}$, $CR_{╕}$ and $CR_{\sim}$.

## 7.3 A truth-value algorithm for TCOI triple implication reasoning

From the nature of logical reasoning, reasoning is a logical process of deriving conclusions from premises. An algorithm for reasoning computes the conclusion by a prescribed procedure rather than deriving it logically.

In triple reasoning, the implication-based reasoning is essentially a semantically driven process that determines the derivability of potential triples according to semantic entailment relations. The algorithmization of reasoning can employ different computational models, including closure computation [50], forward-chaining inference [51], rule entailment checking [52], and truth-value computation under many-valued semantics [53]. From classical logic to modern approaches that use non-classical logics to handle uncertainty, fuzziness and even context dependence, all provide ways to compute the truth value of triples [54, 55].

In this section, we propose a truth value (continuous-value) algorithm for the three fuzzy TCOI triple implication reasoning $T_{\neg}$, $T_{╕}$ and $T_{\sim}$, and perform truth-value calculations for the three fuzzy counterfactual reasoning reasoning $CR_{\neg}$, $CR_{╕}$ and $CR_{\sim}$.

In $T_{\neg}$, $T_{╕}$ and $T_{\sim}$, negation and implication $\rightarrow$ (logical implication) are the core concepts. For the three types of negation in $T_{\neg}$, $T_{╕}$ and $T_{\sim}$, we use the symbol $\mathtt{N}$ to denote these different negations, that is $\mathtt{N} \in \{\neg, ╕, \sim\}$. As for implication $\rightarrow$ in $T_{\neg}$, $T_{╕}$ and $T_{\sim}$, it is necessary to determine $\rightarrow$ itself as well as its relationship with $\mathtt{N}$.

The implication $\rightarrow$ is a binary function in fuzzy logic. We define it as follows with reference to [56] and [57].

**Definition 2**. A mapping $T$: $[0, 1]^2 \rightarrow [0, 1]$ is called a *t-norm*, if it satisfies

(T1) Boundary conditions, $T(0, 0) = 0$, $T(1, x) = x$, $T(0, x) = 0$;

(T2) Monotony, if $x \le y$ then $T(x, z) \le T(y, z)$, for all $z \in [0, 1]$;

(T3) Commutative, $T(x, y) = T(y, x)$, for all $x, y \in [0, 1]$;

(T4) Associative, $T(x, T(y, z)) = T(T(x, y), z)$, for all $x, y, z \in [0, 1]$.

**Definition 3**. Let $T$ be a fixed *t-norm* (e.g., the Gödel *t-norm*, product *t-norm*, or Łukasiewicz *t-norm*). A mapping $I_{COI}$: $[0, 1]^2 \rightarrow [0, 1]$ is called logical implication in fuzzy TCOI triple implication reasoning, if it satisfies

$$I_{COI}(x, y) = sup\{s \in [0,1] \mid T(\mathtt{N}(y), s) \le \mathtt{N}(x)\}, \forall x, y \in [0, 1].$$

According to the above definitions, it is easy to prove that $I_{COI}$ has the following properties:

**Property 5**. For all $x, y \in [0, 1]$,

(a) $T(\mathtt{N}(y), I_{COI}(x, y)) \le \mathtt{N}(x)$;

(b) $T(\mathtt{N}(y), s) \le \mathtt{N}(x)$ if and only if $s \le I_{COI}(x, y)$.

Proof:

(a). Suppose $T(\mathtt{N}(y), I_{COI}(x, y)) > \mathtt{N}(x)$. $\mathtt{N}(y) > \mathtt{N}(x)$ and $I_{COI}(x, y) > \mathtt{N}(x)$ according to Definition 2. Since $I_{COI}(x, y) = sup\{s \in [0,1] \mid T(\mathtt{N}(y), s) \le \mathtt{N}(x)\}$, that is, $T(\mathtt{N}(y), s) \le \mathtt{N}(x)$, it follows that $\mathtt{N}(y) \le \mathtt{N}(x)$. This leads to a contradiction. Therefore, $T(\mathtt{N}(y), I_{COI}(x, y)) \le \mathtt{N}(x)$.

(b). If $T(\mathtt{N}(y), s) \le \mathtt{N}(x)$, then $s = \{s \in [0,1] \mid T(\mathtt{N}(y), s) \le \mathtt{N}(x)\} \le sup\{s \in [0,1] \mid T(\mathtt{N}(y), s) \le \mathtt{N}(x)\} = I_{COI}(x, y)$. If $s \le I_{COI}(x, y)$, then $s \le sup\{s \in [0,1] \mid T(\mathtt{N}(y), s) \le \mathtt{N}(x)\}$. Therefore, $T(\mathtt{N}(y), s) \le \mathtt{N}(x)$. □

For the three fuzzy TCOI triple implication reasoning $T_{\neg}$, $T_{╕}$ and $T_{\sim}$, we propose a truth-value algorithm based on the TCOI-entailment (Definition 1), the $I_{COI}$ implication (Definition 3) and the continuous-valued interpretation $\partial$ of the logic LCOI+PLCOI (Definition 2 in Section 4.2.2).

In accordance with the structure of $T_{\neg}$, $T_{╕}$ and $T_{\sim}$, this algorithm is designed to determine the truth values of the conclusions ¬A, ╕A and ~A.

(I) Algorithm of $CR_{\neg}$

formal expression：A→B, ¬B ╞ ¬A

algorithm: $\partial(\neg A) = T(\partial(\neg B), I_{COI}(\partial(A), \partial(B)))$, $\forall \partial(A), \partial(B) \in [0, 1]$.

(II) Algorithm of $CR_{╕}$

formal expression：A→B, ╕B ╞ ╕A

algorithm: $\partial(╕ A) = T(\partial(╕ B), I_{COI}(\partial(A), \partial(B)))$, $\forall \partial(A), \partial(B) \in [0, 1]]$.

(III) Algorithm of $CR_{\sim}$

formal expression：A→B, ~B ╞ ~A

algorithm: $\partial(\sim A) = T(\partial(\sim B), I_{COI}(\partial(A), \partial(B)))$, $\forall \partial(A), \partial(B) \in [0, 1]$.

Among them, $\partial$(A), $\partial$(¬A), $\partial$(╕A), $\partial$(~A), $\partial$(B), $\partial$(¬B), $\partial$(╕B) and $\partial$(~B) represent the truth values of formulas A, ¬A, ╕A, ~A, B, ¬B, ╕B and ~B in the logic LCOI+PLCOI, respectively.

As can be seen from the structure of Algorithms (I), (II) and (III), the computation of $\partial(\neg A)$, $\partial(╕A)$ and $\partial(\sim A)$ depends on $I_{COI}(\partial(A), \partial(B))$, $\partial(\neg B)$, $\partial(╕B)$ and $\partial(\sim B)$. Therefore, the steps to solve the reasoning conclusions $\partial(\neg A)$, $\partial(╕A)$ and $\partial(\sim A)$ using algorithms (I), (II), and (III) are as follows:

(i) According to the continuous-valued semantics $\partial$ of the logic LCOI+PLCOI, computing the truth value of implication A→B, namely $I_R(\partial(A), \partial(B))$.

(ii) Based on $\partial$ and algorithms (I), (II), (III), computing the truth values $\partial(\neg B)$, $\partial(╕B)$ and $\partial(\sim B)$ of ¬B, ╕B and ~B.

(iii) By (i) and (ii), obtain the truth values $\partial(\neg A)$, $\partial(╕A)$, $\partial(\sim A)$ of the reasoning conclusions ¬A, ╕A and ~A.

Below, we use this procedure to solve the truth values of the reasoning conclusions in the fuzzy counterfactual reasoning $CR_{\neg}$, $CR_{╕}$ and $CR_{\sim}$.

(i) In $CR_{\neg}$, $CR_{╕}$ and $CR_{\sim}$, the first premise is A→B ("If the water temperature in the teacup is high, then the hand feels that the teacup is hot"). By the definition of $\partial$, the truth value $\partial(A\to B)$ of A→B, that is, $I_{COI}(\partial(A), \partial(B)) \in [0, 1]$. The truth values $\partial(\neg B)$, $\partial(╕B)$ and $\partial(\sim B)$ of the second premise ¬B, ╕B and ~B, can be obtained by the truth value $\partial(B)$ of B according to $\partial$.

In fuzzy logic, the truth value of a fuzzy proposition in a domain can be determined through expert opinions or statistical methods. For ease of discussion, we assume $\partial(A\to B) = 0.9$, $\partial(B) = 0.9$.

(ii) Because $\partial(B) = 0.9$, $\partial(╕B) = 1- \partial(B) = 0.1$ by the definition of $\partial$. According to the algorithm (II), $\partial(╕A) = T(\partial(╕B), I_{COI}(\partial(A), \partial(B))) = T(0.1, 0.9) = 0.1$. According to [3] in the definition of $\partial$, $\partial(\sim B)$ has two cases (a) and (d).

For the case (a), $\partial(\sim B) = \lambda - \frac{2\lambda-1}{1-\lambda}(\partial(B)-\lambda)$ when $\lambda\in[½, 1)$ and $\partial(B)\in(\lambda, 1]$, so $\partial(\sim B) = \lambda - \frac{2\lambda-1}{1-\lambda}(0.9-\lambda) = \frac{9-8\lambda}{10(1-\lambda)} - \lambda$. It can be concluded that $\partial(\sim B) < 0.9$. According to the algorithm (III), $\partial(\sim A) = T(\partial(\sim B), I_{COI}(\partial(A), \partial(B))) = T(\partial(\sim B), 0.9) = \partial(\sim B)$. So, $\partial(\sim A) = \partial(\sim B) = \frac{9-8\lambda}{10(1-\lambda)} - \lambda$. Based on [4] in the definition of $\partial$, $\partial(\neg B) = \max(\partial(╕B), \partial(\sim B))$, thus $\partial(\neg B) = \partial(\sim B)$. According to the algorithm (I), $\partial(\neg A) = T(\partial(\neg B), I_{COI}(\partial(A), \partial(B)))$, so $\partial(\neg A) = T(\partial(\sim B), 0.9)$. Thus, $(\neg A) = \partial(\sim B) = \frac{9-8\lambda}{10(1-\lambda)} - \lambda$.

For the case (d), $\partial(\sim B) = 1- \frac{1-2\lambda}{\lambda}(\partial(B)+\lambda-1) - \lambda$ when $\lambda\in(0, ½]$ and $\partial(B)\in(1-\lambda, 1]$, so $\partial(\sim B) = \frac{1-2\lambda}{10\lambda} + \lambda$. It can be concluded that $\partial(\sim B) < 0.9$. According to the algorithm (III), $\partial(\sim A) = T(\partial(\sim B), I_{COI}(\partial(A), \partial(B))) = T(\partial(\sim B), 0.9) = \partial(\sim B)$, so $\partial(\sim A) = \partial(\sim B) = \frac{1-2\lambda}{10\lambda} + \lambda$. Based on [4] in the definition of $\partial$, $\partial(\neg B) = \max(\partial(╕B), \partial(\sim B))$, thus $\partial(\neg B) = \partial(\sim B)$. According to the algorithm (I), $\partial(\neg A) = T(\partial(\neg B), I_{COI}(\partial(A), \partial(B)))$, so $\partial(\neg A) = T(\partial(\sim B), 0.9)$. Thus, $(\neg A) = \partial(\sim B) = \frac{1-2\lambda}{10\lambda} + \lambda$.

(iii) Under the assumption of $\partial(A\to B) = 0.9$, $\partial(B) = 0.9$, based on the above calculation results, the truth values $\partial(\neg A)$, $\partial(╕A)$ and $\partial(\sim A)$ of the conclusions ¬A, ╕A and ~A for the fuzzy counterfactual reasoning $CR_{\neg}$, $CR_{╕}$ and $CR_{\sim}$ as follows:

(1) $\partial(╕A) = 0.1$.

(2) $\partial(\sim A) = \frac{9-8\lambda}{10(1-\lambda)} - \lambda$ ($\lambda\in[½, 1)$), or $\partial(\sim A) = \frac{1-2\lambda}{10\lambda} + \lambda$ ($\lambda\in(0, ½]$).

(3) $\partial(\neg A) = \partial(\sim A)$.

In the above truth values $\partial(\sim A)$, $\lambda$ ($\lambda\in(0, 1)$) is a variable parameter. From the definition of $\partial$ in the continuous-valued semantics of the logic LCOI+PLCOI and Figure 7, it can be seen that the variation in the value of $\lambda$ determines the magnitude and range of ∂(¬A), ∂(╕A) and ∂(~A), that is, $\lambda$ serves as a "threshold" for the range of these truth values. The role and significance of $\lambda$ have been discussed in the context of the set SCOI and the logic LCOI+PLCOI [13].

## 8. Conclusions and future work

Regarding the three different types of negation present in triples and their elements, the classical triple <s, p, o>, which expresses simple positive semantic relationships, as well as classical and non-classical logics that contain only one type of negation (classical negation) in formal languages, both lack inherent mechanisms to distinguish and express these three different types of negation. Therefore, to accurately and finely represent these different negations and their relationships in knowledge representation, the classical triple needs to be extended based on a non classical logic that can distinguish and express different negations at both the syntactic and semantic interpretation levels. Such logic includes multiple negation operators or semantic clarification mechanisms to achieve complete expression of negation meanings and accurate semantic modeling.

Concerning the negativity in triples and their elements, this paper conceptually proposes that there are three distinct negations within triples and their elements: contradictory negation, opposite negation, and intermediary negation. Based on the set SCOI and the logic LCOI+PLCOI, which feature these three negations, an extension of the triple is proposed to distinguish and express the three different negations in triples and their elements, termed the “TCOI triple with contradictory negation, opposite negation and intermediary negation”.

The TCOI triple can express the relationship between the subject and the object along with their degree of association, and it can distinguish and express the three different negations within triples and their elements. This paper demonstrates the expressive power of the TCOI triple through everyday statements and Web resource descriptions involving different negations.

For the TCOI triple reasoning, the focus is primarily on implication-based reasoning. Based on the semantics of logic LCOI+PLCOI, a concept called “TCOI-entailment” is introduced as the semantic implication for TCOI triple implication reasoning. TCOI-entailment establishes a connection between TCOI triple implication reasoning and inferences in logic LCOI+PLCOI, showing that the formal inference laws proven in logic LCOI+PLCOI are valid in TCOI triple implication reasoning. LCOI+PLCOI provide a logical foundation for TCOI triple implication reasoning.

The TCOI triple reasoning and counterfactual reasoning both belong to the category of semantic-based reasoning. Triple reasoning primarily focuses on relational semantics and structured semantics, while counterfactual reasoning mainly emphasizes causal semantics. To demonstrate the reasoning capability of TCOI triple implication reasoning, we apply it to counterfactuals and counterfactual reasoning. Through an example of fuzzy counterfactuals and counterfactual reasoning, it show that the three types of fuzzy counterfactual reasoning based on different negations correspond to three types of fuzzy TCOI triple implication reasoning. Furthermore, we propose a truth-value (continuous-valued) algorithm for fuzzy TCOI triple implication reasoning and perform calculations on the reasoning example.

We consider the TCOI triple is a semantic and structural extension of the classical triple <s, p, o>. While retaining the ability to express affirmative assertions, it systematically introduces the three semantic dimensions of contradictory negation, opposite negation and intermediary negation, allowing these negations to independently apply to the elements (s, p, o) of the triple and on the whole triple. This significantly enhances the triple model capability to represent and reasoning about complex negative information.

Building on this paper, we will further explore the applications of TCOI triple and its reasoning in fields such as RDF, the semantic web and knowledge graphs.

# References

[1] Graham Klyne，Jeremy J. Carroll. Resource Description Framework (RDF): Concepts and Abstract Syntax. http://www.w3.org/TR/2004/REC-rdf-concepts-20040210/

[2] Xiaojun Chen, Shengbin Jia, Yang Xiang, A review: Knowledge reasoning over knowledge graph, Expert Systems with Applications, Volume 141, 2020, 112948

[3] hang, W., Wang, B., Zhu, P., Ding, L., & Wang, S. (2024). A Span-based Multivariate Information-aware Embedding Network for joint relational triplet extraction of threat intelligence. Knowledge-Based Systems, 295, 111829. https://doi.org/10.1016/j.knosys.2024.111829

[4] Marcos, E. do N., & Volkov, Y. (2022). Homogeneous triples for homogeneous algebras with two relations. Journal of Algebra, 599, 1–47. https://doi.org/10.1016/j.jalgebra.2022.01.014

[5] Xie, P., Zhou, G., Liu, J., & Huang, J. X. (2023). Incorporating global–local neighbors with Gaussian mixture embedding for few-shot knowledge graph completion. Expert Systems with Applications, 234, 121086. https://doi.org/10.1016/j.eswa.2023.121086

[6] Qiu, J., Sun, L., & Han, M. (2023). Improving Knowledge Base Updates with CAIA: A Method Utilizing Capsule Network and Attentive Intratriplet Association Features. Journal of Sensors, 2023(1). https://doi.org/10.1155/2023/9942486

[7] Xu, D., Zhu, H., Huang, Y., Jin, Z., Ding, W., Li, H., & Ran, M. (2023). Vision-knowledge fusion model for multi-domain medical report generation. Information Fusion, 97, 101817. https://doi.org/10.1016/j.inffus.2023.101817

[8] Liu, H., Hu, K., Wang, F.-L., & Hao, T. (2020). Aggregating neighborhood information for negative sampling for knowledge graph embedding. Neural Computing and Applications, 32(23), 17637–17653. https://doi.org/10.1007/s00521-020-04940-5

[9] L. R. Horn, H. Wansing. Negation. Stanford Encyclopedia of Philosophy. Zalta, E.N. (ed.). Stanford University, 2020. http://plato. stanford.edu/entries/negation/

[10] K. Makkar *et al*. Improvisation in Opinion Mining Using Negation Detection and Negation Handling Techniques: A Survey. Soft Computing: Theories and Applications. Lecture Notes in Networks and Systems 627, 2023, 799–808.

[11] Morante R, Blanco E. Recent advances in processing negation. Natural Language Engineering, 2021, 27(2): 121-130.

[12] Zhenghua Pan, Yong Wang. Three Kinds of Negation in Knowledge and Their Mathematical Foundations. arXiv:2505.24422 https://doi.org/10.48550/arXiv.2505.24422

[13] Zhenghua Pan, Yong Wang. Different negations and fuzziness in Web resources and resource description and their mathematical foundation, Information Sciences, 689 (2025), 121254. https://doi.org/10.1016/j.ins.2024.121254

[14] Straccia, U., Casini, G. (2022) A Minimal Deductive System for RDFS with Negative Statements. 19th International Conference on Principles of Knowledge Representation and Reasoning, KR 2022. https://doi.org/10.24963/kr.2022/35

[15] Damásio, C., Analyti, A., Antoniou, G. (2010) Embeddings of simple modular extended RDF. Lecture Notes in Computer Science (including subseries Lecture Notes in Artificial Intelligence and Lecture Notes in Bioinformatics). https://doi.org/10.1007/978-3-642-15918-3_17

[16] Damásio, C.V., Analyti, A., Antoniou, G. (2010) Implementing simple modular ERDF ontologies. Frontiers in Artificial Intelligence and Applications. https://doi.org/10.3233/978-1-60750-606-5-1083

[17] Arnaout, H., Razniewski, S., Weikum, G., Pan, J.Z. (2021) Negative statements considered useful. *Journal of Web Semantics*. https://doi.org/10.1016/j.websem.2021.100661

[18] Velios, Athanasios, Meghini, Carlo, Doerr, Martin, Stead, Stephen (2023) Typed properties and negative typed properties: Dealing with type observations and negative statements in the CIDOC CRM. *Semantic Web*. https://doi.org/10.3233/SW-223159

[19] Razniewski, S., Arnaout, H., Ghosh, S., Suchanek, F. (2024) Completeness, Recall, and Negation in Open-world Knowledge Bases: A Survey. *ACM Computing Surveys*. https://doi.org/10.1145/3639563

[20] Angles, R., Gutiérrez, C. (2016) Negation in SPARQL. *CEUR Workshop Proceedings*. https://www.scopus.com/pages/publications/84985961254

[21] Losemann, K., Martens, W. (2012) The complexity of evaluating path expressions in SPARQL. *Proceedings of the ACM SIGACT-SIGMOD-SIGART Symposium on Principles of Database Systems*. https://doi.org/10.1145/2213556.2213573

[22] Losemann, K., Martens, W. (2013) The Complexity of regular expressions and property paths in sparql. *ACM Transactions on Database Systems*. https://doi.org/10.1145/2494529

[23] Darari, F., Nutt, W., Razniewski, S., Rudolph, S. (2020) Completeness and soundness guarantees for conjunctive SPARQL queries over RDF data sources with completeness statements. *Semantic Web*. https://doi.org/10.3233/SW-190344

[24] Straccia, U., Casini, G. (2022) A Minimal Deductive System for RDFS with Negative Statements. *19th International Conference on Principles of Knowledge Representation and Reasoning, KR 2022*. https://doi.org/10.24963/kr.2022/35

[25] Darari, F. (2013) Representing and querying negative knowledge in RDF. *Lecture Notes in Computer Science (including subseries Lecture Notes in Artificial Intelligence and Lecture Notes in Bioinformatics)*. https://doi.org/10.1007/978-3-642-41242-4_40

[26] Alkhateeb, F. (2016) Introducing wild-card and negation for optimizing SPARQL queries based on rewriting RDF graph and SPARQL queries. *WEBIST 2016 - Proceedings of the 12th International Conference on Web Information Systems and Technologies*. https://doi.org/10.5220/0005762001810187

[27] Arnaout, H., Razniewski, S. (2023) Can large language models generate salient negative statements?. *CEUR Workshop Proceedings*. https://www.scopus.com/pages/publications/85179558782

[28] K. Kangal. Engels' Intentions in Dialectics of Nature. Science & Society, 2019, 83(2):215-243. DOI: 10.1521/siso.2019.83.2.215

[29] Mazzieri M, Dragoni A F. A fuzzy semantics for the resource description framework. URSW 2005–2007. 2008, 244–261

[30] Hayes, P. RDF Semantics. W3C Recommendation, 2004

[31] Schneider, M. A survey of RDF data management. The VLDB Journal, 2009, 19(1): 1-22

[32] Bordes, A., et al. Translating embeddings for modeling multi-relational data. Advances in Neural Information Processing Systems 26 (NeurIPS). 2013: 2787-2795

[33] Trouillon, T., et all. Complex embeddings for simple link prediction. Proceedings of the 33rd International Conference on Machine Learning (ICML). New York: PMLR, 2016: 2071-2080

[34] Schlichtkrull, M., et al. Modeling relational data with graph convolutional networks. Proceedings of the Extended Semantic Web Conference (ESWC). Cham: Springer, 2018: 593-607

[35] Yang, F., et al. Differentiable learning of logical rules for knowledge base reasoning. Advances in Neural Information Processing Systems 30 (NeurIPS). 2017: 2319-2328

[36] Lao, N., Cohen, W. W. Relational retrieval using a combination of path-constrained random walks. Machine Learning, 2010/2011.

[37] Nickel M, Murphy K, Tresp V, et al. A review of relational machine learning for knowledge graphs. Proceedings of the IEEE, 2016, 104(1): 11-33. DOI: 10.1109/JPROC.2015.2483592

[38] Galárraga, L., et al. AMIE: association rule mining under incomplete evidence in ontological knowledge bases. Proceedings of the 22nd International Conference on World Wide Web (WWW 2013). New York: ACM, 2013: 413-422

[39] Galárraga, L., et al. Fast rule mining in ontological knowledge bases with AMIE+. The VLDB Journal, 2015, 24(5): 707-730. DOI:10.1007/s00778-015-0394-1

[40] Ter Horst H J. Completeness, decidability and complexity of entailment for RDF Schema and a semantic extension involving the OWL vocabulary. Journal of Web Semantics, 2005, 3(1): 71-110

[41] Sadeghian, A, et al. DRUM: end-to-end differentiable rule mining on knowledge graphs. Advances in Neural Information Processing Systems 32 (NeurIPS 2019). Red Hook: Curran Associates, 2019

[42] Lewis, David K. (1973). *Counterfactuals*. Cambridge, MA: Harvard University Press

[43] Roese, N. J. (1997). Counterfactual thinking. *Psychological bulletin*, *121*(1), 133

[44] Edgington, D. (2020). Counterfactual Conditionals. In *The Routledge Handbook of Modality* (pp. 30–39). Routledge. https://doi.org/10.4324/9781315742144-5

[45] Vorster, E. (2022). Does Counterfactual Reasoning Hold the Key to Artificial General Intelligence?. In: Pillay, A., Jembere, E.,

Gerber, A. (eds) Artificial Intelligence Research. SACAIR 2022. Communications in Computer and Information Science, vol 1734. Springer, Cham. https://doi.org/10.1007/978-3-031-22321-1_26

[46] Fan, D. (2023). Focused true–true counterfactuals. *The Philosophical Forum*, *54*(3), 121–141. https://doi.org/10.1111/phil.12337

[47] Kühl, C. E. (2023). On Counterfactual Reasoning. Danish Yearbook of Philosophy, 56(2), 154–181. https://doi.org/10.1163/24689300-bja10043

[48] Chou, Y. L., Moreira, C., Bruza, P., Ouyang, C., & Jorge, J. (2022). Counterfactuals and causability in explainable artificial intelligence: Theory, algorithms, and applications. Information Fusion, 81, 59-83. https://doi.org/10.1016/j.inffus.2021.11.003

[49] Cerami M , Pardo P . Many-Valued Semantics for Vague Counterfactuals. Studies in Logic, 36, 341-362 (2011)

[50] Ganter, B., & Wille, R. (1999). Formal Concept Analysis: Mathematical Foundations. Springer.

[51] Matheus, C. J., et al. 2006. BaseVISor: A Triples-Based Inference Engine Outfitted to Process RuleML and R-Entailment Rules. https://doi.org/10.21236/ada460530

[52] Y Hogan A, Blomqvist E, Cochez M, et al. Knowledge graphs. ACM Computing Surveys, 2021, 54(4): 1-37

[53] Liu, Chang et al. Fuzzy Reasoning over RDF Data Using OWL Vocabulary. 2011 IEEE/WIC/ACM International Conferences on Web Intelligence and Intelligent Agent Technology 1 (2011): 162-169.

[54] Xiaojun Chen, Shengbin Jia, Yang Xiang. A review: Knowledge reasoning over knowledge graph. Expert Systems With Applications, 2020, 141: 1–21.

[55] Xinliang Liu, Tingyu Mao, et al. Overview of knowledge reasoning for knowledge graph. Neurocomputing, Volume 585, 2024, 127571. https://doi.org/10.1016/j.neucom.2024.127571

[56] Bustince, H., Burillo, P.& Soria, F. Automorphisms, negations and implication operators. Fuzzy Sets & Systems, 2003, 134(2): 209–229

[57] Aguiló, I., Massanet, S., Riera, J.V., Ruiz-Aguilera, D. Modus Ponens Tollens for RU-Implications. In: Lesot, MJ., et al. Information Processing and Management of Uncertainty in Knowledge-Based Systems. IPMU 2020. Communications in Computer and Information Science, vol 1238, 2020. Springer, Cham. https://doi.org/10.1007/978-3-030-50143-3_61